\documentclass[12pt]{article}
\usepackage[english]{babel} 
\usepackage{fullpage,graphicx,amsmath,amsthm,amsfonts,verbatim}
\usepackage[small,bf]{caption}
\usepackage{tikz} 
\usepackage{pgf,pgfplots}
\usepackage{import}
\usepackage{url}
\usepackage[colorlinks=true]{hyperref}
\def\layersep{2.5cm}

\newcommand{\ones}{\mathbf 1}
\newcommand{\reals}{{\mbox{\bf R}}}
\newcommand{\mask}{{\mathcal M}}

\newcommand{\prox}{\mathbf{prox}}
\newcommand{\mprox}{\mathbf{mprox}}
\newcommand{\wprox}{\mathbf{wprox}}

\newcommand{\eqK}{\mathop{\stackrel{\mathcal K}{=}}}

\newcommand{\Prob}{\mathop{\bf Prob}}

\newcommand{\argmin}{\mathop{\rm argmin}}

\newcommand{\eg}{{\it e.g.}}
\newcommand{\ie}{{\it i.e.}}

\newcommand{\BEAS}{\begin{eqnarray*}}
\newcommand{\EEAS}{\end{eqnarray*}}
\newcommand{\BEA}{\begin{eqnarray}}
\newcommand{\EEA}{\end{eqnarray}}
\newcommand{\BEQ}{\begin{equation}}
\newcommand{\EEQ}{\end{equation}}
\newcommand{\BIT}{\begin{itemize}}
\newcommand{\EIT}{\end{itemize}}

\renewcommand{\hat}{\widehat}
\renewcommand{\tilde}{\widetilde}

\newcounter{algorithmctr}[section]
\renewcommand{\thealgorithmctr}{\thesection.\arabic{algorithmctr}}
\newenvironment{algdesc}%
   {\refstepcounter{algorithmctr}\begin{list}{}{%
       \setlength{\rightmargin}{0\linewidth}%
       \setlength{\leftmargin}{.05\linewidth}}%
       \rmfamily\small
       \item[]{\setlength{\parskip}{0ex}\hrulefill\par%
        \nopagebreak{\bfseries\textsf{Algorithm \thealgorithmctr~}}}}%
   {{\setlength{\parskip}{-1ex}\nopagebreak\par\hrulefill} \end{list}}

\title{Signal Decomposition\\ Using Masked Proximal Operators}
\author{Bennet E. Meyers \and Stephen P. Boyd}

\begin{document}
	
\maketitle

\begin{abstract}
We consider the well-studied problem of decomposing a vector time series signal
into components with different characteristics, such as smooth, periodic, 
nonnegative, or sparse.
We describe a simple and general framework in which the components
are defined by loss functions (which include constraints), 
and the signal decomposition is carried out by minimizing the sum of losses of 
the components (subject to the constraints). When each loss function is the 
negative log-likelihood of a density for the signal component, this framework 
coincides with maximum a posteriori probability (MAP) estimation; 
but it also includes many other interesting cases.
Summarizing and clarifying prior results, we give two distributed optimization 
methods for computing the decomposition, which find
the optimal decomposition when the component class loss functions
are convex, and are good heuristics when they are not.
Both methods require only the masked proximal operator of 
each of the component loss functions, a generalization of
the well-known proximal operator that handles missing entries 
in its argument.
Both methods are distributed, \ie, handle each component separately.
We derive tractable methods for evaluating the masked proximal 
operators of some loss functions that, to our knowledge, have 
not appeared in the literature.
\end{abstract}

\clearpage
\tableofcontents
\clearpage

\section{Introduction}

The decomposition of a time series signal into components is an age 
old problem, with many different approaches proposed, including traditional 
filtering and smoothing, seasonal-trend decomposition, 
Fourier and other decompositions,
principal component analysis (PCA),
and newer variants such as nonnegative matrix factorization, 
various statistical methods, and many heuristic methods.
It is believed that ancient Babylonian mathematicians used harmonic 
analysis to understand astronomical observations as collections 
of `periodic phenomena'~\cite{neugebauer1969}.

As we will discuss in detail in \S\ref{s-related-work}, 
formulating the problem of decomposing a time series signal 
into components as an optimization problem has a long history. We introduce a 
simple framework that 
unifies many existing approaches, where components are described by their loss
functions.
Once the component class loss functions are chosen, we minimize
the total loss subject to replicating the given signal with the 
components.
We give a simple unified algorithm, based on variations of well-known 
algorithms, for carrying out this 
decomposition, which is guaranteed to find the globally optimal
decomposition when the loss functions are all convex, and
is a good heuristic when they are not.
The method accesses the component loss functions only through a 
modified proximal operator interface, which takes into account that some
data in the original signal may be missing.
The method is distributed, in that each component class is handled 
separately, with the algorithm coordinating them.

\paragraph{Handling of missing data.}
The methods discussed in this paper are designed to handle missing data in the 
original 
signal to be decomposed, a common situation in many practical settings.
The signal components in the decomposition, however, do not have
any missing data;
by summing the components in the decomposition, we obtain a 
guess or estimate of the missing values in the original signal.
This means that 
signal decomposition can be used as a sophisticated method
for guessing or imputing or interpolating missing or unknown entries in a 
signal.
This allows us to carry out 
a kind of validation or self-consistency check on a decomposition,
by pretending that some known entries are missing, and comparing the 
imputed values to the known ones.

\paragraph{Expressivity and interpretability.}
The general framework described here includes many well-known problems as 
specific instances, and it enables the design of newer, more complex components 
classes than traditional simple ones
such as a periodic signal, a trend, a smooth signal, and so on.  For example we 
can define a signal component class that consists of periodic, smooth, and 
nonnegative
signals, or piecewise constant signals that have no more than some specified
number of jumps.
The resulting decomposition is always interpretable, since we specify the
component classes.

\paragraph{Outline.}
We describe the signal decomposition framework in \S\ref{s-SD}, 
where we pose signal decomposition as an optimization problem,
concluding with an illustrative simple example in~$\S$\ref{s-simple-example}.
In \S\ref{s-related-work} we cover related and previous work and methods.
Two distributed methods for solving the signal decomposition problem, based on 
variations of well established algorithms, 
are described in \S\ref{s-solution-method}.
The next two sections concern loss functions for signal component classes:
general attributes are described in \S\ref{s-class-attr} and
some example classes in \S\ref{s-classes}. The topic of how to fit
component class losses given archetypal examples is discussed in
\S\ref{s-fitting-losses}.
We conclude the monograph with examples using real data:
Weekly CO2 measurements at Mauna Loa in \S\ref{s-CO2},
hourly traffic over a New York bridge in \S\ref{s-traffic},
and 1-minute power output for a group (fleet) of seven photo-voltaic (PV)
installations in \S\ref{s-pv-fleet}.

\paragraph{Software.}
Our paper is accompanied by an open-source  software implementation 
called \texttt{OSD}, short for `Optimization(-based) Signal Decomposition',
available at 
\begin{quote}
\url{https://github.com/cvxgrp/signal-decomposition}.
\end{quote}

\section{Signal decomposition}\label{s-SD}
\subsection{Signal decomposition into components}
\paragraph{Vector time series signal with missing entries.}
Consider a vector time series or signal, possibly with missing entries,
$y_1, \ldots, y_T \in (\reals \cup \{?\})^p$.  We denote the $i$th entry of 
$y_t$
as $(y_t)_i = y_{t,i}$.
The value $?$ denotes a missing entry in the signal; we say that entry 
$y_{t,i}$ is known if $y_{t,i}\in \reals$, and unknown if $y_{t,i} = ?$.
We define $\mathcal K$ as the set of indices corresponding to known values,
\ie, $\mathcal K = \{(t,i) \mid y_{t,i} \in \reals \}$.
We define $\mathcal U$ as the set of indices corresponding to unknown or 
missing values, \ie, $\mathcal U = \{(t,i) \mid y_{t,i} = ? \}$.
We represent the signal compactly as a $T\times p$ matrix
$y \in (\reals \cup \{?\})^{T \times p}$, with rows $y_1^T, \ldots, y_T^T$.

\paragraph{The mask operator.}  Let $q = |\mathcal K|$ be the total
number of known entries in $y$, with $q \leq Tp$.
We introduce the mask operator $\mask: (\reals \cup \{?\})^{T \times p} \to 
\reals^q$,
which simply lists the entries of its argument that are in $\mathcal K$
in a vector, in some known order.
We will also use its adjoint $\mask^*$, which takes a vector in $\reals^q$
and puts them into a $T \times p$ matrix, in the correct order, with other
entries zero.
Note that while the original signal $y$ can have missing entries, 
the vector $\mask y$ does not.
We also observe that for any $z \in \reals^{T\times p}$, 
$\mask^* \mask z$ is $z$, with the entries in $\mathcal U$ replaced with zeros.

\paragraph{Signal decomposition.}
We will model the given signal $y$ as a sum (or decomposition) of $K$ components
$x^1, \ldots, x^K \in \reals^{T \times p}$,
\[
y_{t,i} = (x^1)_{t,i}+\cdots+(x^K)_{t,i}, \quad (t,i)\in \mathcal K.
\]
We refer to this constraint, that the sum of the components matches the 
given signal at 
its known values, as the consistency constraint, which can be expressed as
\BEQ\label{e-decomp}
\mask y = \mask x^1+\cdots+ \mask x^K.
\EEQ

Note that the components $x^1, \ldots, x^K$ do not have missing values.
Indeed, we can interpret the values
\BEQ\label{e-missing-est}
\hat y_{t,i} = x^1_{t,i} +\cdots+x^K_{t,i}, \quad
(t,i) \in \mathcal U,
\EEQ
as estimates of the missing values in the original signal $y$.
(This will be the basis of a validation method described later.)

\subsection{Component classes}
The $K$ components are characterized by functions 
$\phi_k: \reals^{T \times p}\to \reals \cup\{\infty\}$, $k=1, \ldots, K$.
We interpret $\phi_k(x)$ as the loss of or implausibility that $x^k=x$.  
We will see later that in some cases we can interpret the classes 
statistically, with $\phi_k(x)$ the negative log-likelihood of $x$ for signal
class $k$.
Roughly speaking, the smaller $\phi_k(x)$ is, the more plausible it is.
Infinite values of $\phi_k(x)$ are used to encode constraints on 
components.
We refer to $x$ as feasible for component class $k$ if 
$\phi_k(x) < \infty$, and we 
refer to $\{x \mid \phi_k(x) < \infty \}$ as the set of feasible signals
for component class $k$. 
When a component class takes on the value $\infty$ for some $x$,
we say that it contains or encodes constraints; when $\phi_k$ does
not take on the value $\infty$, we say the component class has 
no constraints, or has full domain.
We will assume that every component class has at least one feasible 
point, \ie, a point with finite loss.

We will see many examples of component class losses later, but for now we 
mention a few simple examples.

\paragraph{Mean-square small class.}
One simple component class has the mean-square loss
\BEQ\label{e-ms-small}
\phi(x) = \frac{1}{Tp} \sum_{t,i} (x_{t,i})^2 = 
\frac{1}{Tp} \|x\|_F^2,
\EEQ
where $\| \cdot \|_F$ denotes the Frobenius norm, the squareroot of the sum of
squares of the entries.
(To lighten the notation, we drop the subscript $k$ when describing a 
general component class.)
All signals are feasible for this class; roughly speaking, smaller signals 
are more plausible than larger signals.
We call this the component class of mean-square small signals.

We will assume that the first class is always mean-square small, with
loss function \eqref{e-ms-small}.  We interpret $x^1$ 
as a residual in the approximation
\[
y \approx x^2+\cdots+x^K,
\]
and $\phi_1(x^1)$ as the mean-square error.

\paragraph{Mean-square smooth class.}
The component class of mean-square smooth signals has loss
\BEQ\label{e-ms-smooth}
\phi(x) = \frac{1}{(T-1)p} 
\sum_{t=1}^{T-1} \|x_{t+1}-x_t\|_F^2,
\EEQ
the mean-square value of the first difference.  Here too all signals
are feasible, but smooth ones, \ie, ones with small mean-square first 
difference,
are more plausible.

\paragraph{Boolean signal class.}
As one more simple example, consider the component class with loss function
\BEQ\label{e-boolean}
\phi(x) = \left\{ \begin{array}{ll} 0 & x_{t,i} \in \{0,1\}~\mbox{for all}~t,i\\
\infty & \mbox{otherwise}. \end{array} \right.
\EEQ
This component class consists only of constraints, specifically that
each entry is either $0$ or $1$.   It has a finite number, $2^{Tp}$, 
of feasible signals, with no
difference in plausibility among them.  We refer to this class as 
the Boolean component class.

\subsection{Signal decomposition problem}
We will estimate the components $x^1, \ldots, x^K$ by solving 
the optimization problem
\BEQ\label{e-sd}
\begin{array}{ll}
\mbox{minimize}   &  \phi_1(x^1) + \cdots +
\phi_K(x^K)\\
\mbox{subject to} & \mask y  = \mask x^1+ \cdots + \mask x^K,
\end{array}
\EEQ
with variables $x^1, \ldots, x^K$.
We refer to this problem as the \emph{signal decomposition (SD) problem}.
Roughly speaking, we decompose the given signal $y$ into components so as to 
minimize the total implausibility.


We observe that the entries of the mean-square small component 
$x^1$ with indices in $\mathcal U$
do not appear in the contraints, so their optimal value is zero, \ie,
$x^1 = \mask^* \mask x^1$.  It follows that
$\phi_1(x^1) = \frac{1}{Tp} \|\mask x^1 \|_2^2$.
We can now eliminate $x^1$,
and express the SD problem as the unconstrained problem
\BEQ\label{e-sd-no-x1}
\begin{array}{ll}
\mbox{minimize}   & \frac{1}{Tp} \left\| \mask y - \mask x^2 - \cdots - \mask 
x^K 
\right\|_2^2 + \phi_2(x^2) + \cdots + \phi_K(x^K),
\end{array}
\EEQ
with variables $x^2, \ldots, x^K$.
From a solution of this problem we can recover an optimal $x^1$ for
\eqref{e-sd} from the residual in the first term, as
$x^1 =\mask^* (\mask y - \mask x^2 - \cdots - \mask x^K)$.

\paragraph{Solving the signal decomposition problem.}
If the class losses $\phi_k$ are all convex functions, the SD problem 
\eqref{e-sd}
is convex, and can be efficiently solved globally \cite{convex_opt}.
In other cases it can be very hard to find a globally optimal solution, and we
settle for an approximate solution.
In \S \ref{s-solution-method} we will describe two methods that solve the SD 
problem
when it is convex (and has a solution), and approximately solve it when it is 
not.
The first method is based on block coordinate descent (BCD) 
\cite{Beck2013,Wright2015}, and the 
second is based on ADMM \cite{Boyd2011}, an operator splitting method.
Both methods handle each of the component classes separately,
using the masked proximal operators of the loss functions (described
in \S\ref{s-prox}).
This gives a very convenient software architecture, and makes it easy to 
modify or extend it to many component classes.

\paragraph{Existence and uniqueness of decomposition.}
With the assumption that the first component class is mean-square small, and 
all other component
classes contain at least one signal with finite loss, the SD
problem is always feasible.
But it need not have a solution, or when it does, a unique solution.
For example, consider $K=2$ with a mean-square small component and a 
Boolean component.
If the $(t,i)$ entry in $y$ is unknown, then $x^2_{t,i}$ can be either $0$ or 
$1$,
without affecting feasibility or the objective. The uniqueness of specific 
instances of the SD problem (particularly when $K=2$) has been studied 
extensively~\cite{McCoy2014a,Donoho2001a}. (See \S\ref{s-related-work} for a 
longer discussion.)

\subsection{Statistical interpretation}\label{s-stat-interp}
We can give the losses a simple statistical interpretation in some cases,
which conversely can be used to suggest class losses.
Suppose that $\phi$ is continuous on its domain, with 
\[
Z = \int \exp - \phi(x) \; d x < \infty. 
\]
(The integration is with respect to Lebesgue measure.)
We associate with this component class the density
\[
p(x) = \frac{1}{Z} \exp - \phi(x).
\]
Thus, $\phi(x)$ is a constant plus 
the negative log-likelihood of $x$ under this density,
a standard statistical measure of implausibility.
Convex loss functions correspond to log-concave densities.

As an example,
with the mean-square loss $\phi(x) = \frac{1}{2Tp}\|x\|_F^2$ (note the 
additional
factor of two in the denominator), the associated 
density is Gaussian, with the entries of $x$ IID $\mathcal N(0,1)$.
As another example, the mean-square smooth component class with loss 
\eqref{e-ms-smooth}
has $Z= \infty$, so we cannot associate it with a density. 

When all component classes have $Z < \infty$, we can 
interpret the SD problem statistically.
Suppose $x^1, \ldots, x^K$ are independent random variables with
densities $p_1, \ldots, p_K$.
Then the SD objective is a constant plus the negative log-likelihood of 
the decomposition with
$x^1, \ldots, x^K$, and the SD decomposition is the maximum a posteriori 
probability
(MAP) decomposition of the observed signal $y$.


\subsection{Optimality and stationarity conditions}
Here we give optimality or stationarity conditions for the SD problem 
for some special but common cases.
In all cases, the conditions include primal feasibility \eqref{e-decomp},
\ie, consistency,
and a second condition, dual feasibility, which has a form that depends on the 
properties of the losses.

\paragraph{Differentiable losses.}
We first suppose that the losses are differentiable.
The dual feasibility condition is that there exists a Lagrange multiplier 
$\nu \in \reals^q$ for which
\[
\nabla\phi_k(x^k) = \mask^*\nu, \quad k=1, \ldots, K,
\]
where $\nu \in \reals^q$ is a dual variable or 
Lagrange multiplier associated with
the consistency constraint \eqref{e-decomp}.
In words: the gradients of the losses all agree,
and are zero in the unknown entries.
If all losses are convex, this condition together with 
primal feasibility are the necessary 
and sufficient optimality conditions for the SD problem.   If the losses
are not all convex, then this condition together with primal feasibility
are stationarity conditions;
they hold for any optimal decomposition, but
there can be non-optimal points that also satisfy them.

Since $\phi_1(x) = \frac{1}{Tp} \|x\|_F^2$, we have $\nabla \phi^1(x) = 
(2/Tp)x$.
The dual conditions can then be written as
\BEQ\label{e-dual-feas-x1}
x^1 = \mask^*\mask x^1, \qquad \nabla\phi_k(x^k)  = \frac{2}{Tp} x^1,
\quad k=2, \ldots, K,
\EEQ
\ie, the gradients of the component class losses all equal
the mean-square residual, scaled by $2/(Tp)$ in the known entries,
and are zero in the unknown entries.
These are also the conditions under which the gradients of the 
objective in the unconstrained SD problem
formulation \eqref{e-sd-no-x1} with respect to $x^2, \ldots, x^K$ are 
all zero.

\paragraph{Convex losses.}
If the losses are convex but not differentiable, we replace
the gradients in \eqref{e-dual-feas-x1}
with subgradients, to obtain
\BEQ \label{e-dual-feas-subg}
x^1 = \mask^*\mask x^1, \qquad  
g^k = \frac{2}{Tp} x^1, \quad
g^k \in \partial \phi_k(x^k), \quad k=2, \ldots, K,
\EEQ
where $\partial \phi_k(x^k)$ is the subdifferential of $\phi_k$ at $x^k$.
This condition, together with primal feasibility, are optimality conditions
for the SD problem.

\paragraph{Other cases.}
When the losses are neither convex nor differentiable,
the stationarity conditions can be very complex, with the gradients in 
\eqref{e-dual-feas-x1} or 
subgradients in \eqref{e-dual-feas-subg} substituted with some appropriate
generalized gradients.


\subsection{Signal class parameters}
The component class losses $\phi_k$ can also have parameters associated with 
them.
When we need to refer to the parameters, we write $\phi_k(x^k)$ as 
$\phi_k(x^k; \theta_k)$, where $\theta_k \in \Theta_k$, the set of allowable 
parameters.
These parameters are fixed whenever we solve the SD problem, but it is common
to solve the SD problem for several values of the parameters, and choose one
that works well (\eg, using a validation method described later).
The role of the parameters $\theta_k$ will be made clear when we 
look at examples.  For now, though, we mention a few common examples.

\paragraph{Weight or scaling parameters.}
It is very common for a parameter to scale a fixed function, \ie,
$\phi(x; \theta) = \theta \ell(x)$, $\theta \in \Theta = \reals_{++}$,
the set of positive numbers.
(Of course we can have additional parameters as well.)
In this case we interpret the parameters as weights that scale the 
relative implausibility of the component classes.
We will use the more traditional symbol $\lambda$ to denote 
scale factors in loss functions, with the understanding that they are
part of the parameter $\theta$.

\paragraph{Value and constraint parameters.}
Parameters are often used to specify constant values that appear in the loss 
function.
For example we can generalize the Boolean loss function,
which constrains the entries of $x$ to take on values in $\{0,1\}$,
to one where the entries of $x$ take on values in a finite set
$\{\theta_1, \ldots, \theta_M\}$, where $\theta_i \in \reals$, \ie,
\BEQ\label{e-finite-set}
\phi(x) = \left\{ \begin{array}{ll} 0 & 
x_{t,i} \in \{\theta_1, \ldots, \theta_M\}~\mbox{for all} ~t,i\\
\infty & \mbox{otherwise.}
\end{array} \right.
\EEQ
In this case, the parameters give the values that the entries of $x$ are 
allowed to take on.
As another example, consider a loss function that constrains
the entries of $x$ to lie in the interval $[\theta_1, \theta_2]$ 
(with $\theta_1 \leq \theta_2$).  Here the parameters set the lower and upper 
limits on the entries of $x$.

\paragraph{Basis.}
Another common use of parameters is to specify a basis for the 
component, as in
\BEQ \label{e-comp-basis}
\phi(x) = \left\{ \begin{array}{ll} 0 &
x = \theta z ~\mbox{for some}~ z \in \reals^{d \times p}\\
\infty & \mbox{otherwise},
\end{array} \right.
\EEQ
where $\theta \in \reals^{T \times d}$, and $z\in \reals^{d\times p}$.
This component class requires
each column of $x$, \ie, the scalar time series associated with an
entry of $x$,
to be a linear combination of the basis 
(scalar) signals given by the columns of $\theta$ (sometimes referred to 
as a dictionary).
The entries of $z$ give the coefficients of the linear combinations;
for example, the first column of $x$ is
$z_{11}\theta_1 + \cdots + z_{1d} \theta_d$, where 
$\theta_i$ is the $i$th column of $\theta$, \ie, the $i$th
basis signal.

\subsection{Model selection}\label{s-validation}
We refer to a particular choice of component classes and their 
parameter values as an SD model. 
A natural question is: How should we choose the SD model?
In some contexts such as prediction in machine learning the analogous question
of what prediction model we should use, and what parameters we should 
select, has a straightforward answer: We should use the 
model that has the best out-of-sample prediction performance.
(In some cases there are secondary objectives such as model simplicity
or interpretability.)
At the other extreme we have unsupervised machine learning methods, 
such as clustering methods,  where it is 
harder to identify a measure of model performance, and therefore 
harder to find a method for choosing one model over another.
In such cases the model and parameter values are chosen so that the
results correspond to what the user expects or wants to see.
If the model can handle missing data, it can also be checked for 
internal consistency by checking how it imputes values that we actually
know, but pretend while building the model are unknown.
Signal decomposition lies closer to the unsupervised learning setting.

The methods described in this paper are typically 
applied in situations where the analyst has a strong prior belief 
about what they want from a decomposition, often drawn from domain expertise. 
The analyst has a rough sense of the number of components they are 
looking for and the general characteristics of those components, which inform 
the selection of $K$, $\phi_k$, and $\theta_k$.

The classic example of this style of analysis is seasonal-trend 
decomposition (see \S \ref{s-related-work} and \S\ref{s-CO2}), in which a 
scalar signal ($p=1$) 
is decomposed into $K=3$ components: seasonal, trend, and residual. 
(We will see later that this can be approached as an SD problem.)
Here we use the strong prior that the seasonal component should vary smoothly 
over the year, and the trend component must change slowly.
So $K=3$, and the specific forms of the component losses, are not arbitrary;
each has a specific meaning.  In this case the weights or parameters in
the loss functions are chosen to give a plausible or useful 
decomposition.

This can be contrasted and compared with PCA, where we need to determine
the number of principal components $K$ to use.
Aside from the general idea that smaller $K$ is to be preferred over
larger $K$, there is no particular meaning to prefer $K=3$.
In PCA we let the data determine $K$, typically by
finding the smallest $K$ for which the model is at reasonably
self-consistent.

For SD, the specific components, and the form of the loss functions,
are specified by the analyst.  The quality of the decomposition is 
judged using the analyst's domain expertise and intuition.
It is also possible to validate an SD model, or at least, check its 
consistency.  We describe this now.

\paragraph{Model validation.}
We can validate, or at least check consistency of,
a choice of the component classes and their parameter values. 
To do this, we select (typically randomly) some entries of $y$ that are
known, 
denoted $\mathcal T \subset \mathcal K$ (for `test'),
and replace them with the value $?$.  A typical choice of the number of test
entries is a fraction of the known entries, such as $20\%$.
We then carry out the decomposition
by solving the SD problem, using the entries $\mathcal K \setminus \mathcal T$ 
of $y$.
This decomposition gives us estimates or guesses of the entries of $y$
in $\mathcal T$, given by \eqref{e-missing-est}.
Finally, we check these estimates against the true values of $y$,
for example by evaluating the mean-square test error 
\[
\frac{1}{|\mathcal T|p} \sum_{(t,i)\in \mathcal T}
(y_{t,i} - \hat y_{t,i})^2.
\]
A more stable estimate of test error
can be found by evaluating the mean-square test error 
for multiple test sets $\mathcal T^{(1)}, \ldots,\mathcal
T^{(M)}$, each with the same number of entries,
and averaging these to obtain a final mean-square error.

It is reasonable to prefer a model (\ie, choice of component classes
and their parameters) that results in small test error, compared
to another model with higher test error.
The out-of-sample validation method described above can be used to guide the 
choice of the component classes and parameters that define an SD model.

\paragraph{Validating with non-unique decompositions.}
We note that the basic validation method fails when the 
SD problem has multiple solutions, or more precisely, when 
multiple optimal signal decompositions correspond to different values
of $\hat y_{t,i}$ for $(t,i)\in \mathcal T$.
One simple work-around is
to regard the multiple solutions as each providing an
estimate of the missing entry, and to evaluate the test loss
using the best of these estimates.
For example, suppose the second
component class is Boolean, so $(x^2)_{t,i}$ can have the 
value $0$ or $1$ for $(t,i) \in \mathcal T$.
We judge the error using 
\[
\frac{1}{|\mathcal T|p} \sum_{(t,i)\in \mathcal T}
\min_{x^2_{t,i} \in \{0,1\}}
(y_{t,i}-\hat y_{t,i})^2.
\]

\paragraph{Parameter search.} 
As is standard in machine learning and data fitting, it is common to carry out 
multiple decompositions with the same loss functions but different parameters,
and validate each of these choices on one or more test sets, 
as described above.
We then choose as the final parameter values ones corresponding to the lowest 
achieved test error.
As in machine learning and data fitting, the final decomposition is
then fit with all known data, using the parameter values found in
the parameter search.

\subsection{Data pre-processing}
As in other data processing problems, pre-processing the raw data is often 
useful,
leading to better results or interpretability.

\paragraph{Standarization.}
The most basic pre-processing is to standardize the entries of $y$, 
with a scale and offset for each component that results in the average value 
being around zero and the standard deviation around one.
In some cases, for example when the entries of $y$ are all measured in the same
physical units, it can be more appropriate to use 
the same scaling for all components of $y$.

\paragraph{Log transform.}
If the data are all positive and vary over a large range of values, 
a log transform of the raw data can be appropriate.  Roughly speaking, this 
means
that we care about relative or fractional deviations, as opposed to absolute 
errors 
in the raw data, \eg, we consider the values $10$ and $11$ to be as close as 
the 
values $1000$ and $1100$.
With a log transform, the signal decomposition has an interpretation 
as a multiplicative decomposition (in the raw data), as opposed to an additive 
decomposition.  If we denote the raw data as $\tilde y$ and the transformed data
as $y = \log \tilde y$ (entrywise), and the decomposition is $y \eqK  x^1+ 
\cdots +x^K$,
in terms of the raw data we have
\[
\tilde y_{t,i} = \tilde x^1_{t,i} \cdots \tilde x^K_{t,i},
\quad (t,i) \in \mathcal K,
\]
where $\tilde x^i = \exp x^i$ (entrywise), $i=1, \ldots, K$.
The signals $\tilde x^i$ can be though of as multiplicative factors.

\subsection{Simple example}\label{s-simple-example}
In this section we give a simple synthetic example to illustrate
the idea.

\paragraph{Signal decomposition model.}
We construct an SD problem with $p=1$ (\ie, a scalar signal), 
$T=500$, and $K=3$ component classes:
mean-square small, mean-square second-order smooth, and a scaled Boolean.
For mean-square small we use loss function \eqref{e-ms-small},
and for mean-square second order smooth we use the loss
\BEQ\label{e-ss-second}
\phi_2(x) = \frac{\theta_1}{(T-2)p} \sum_{t=2}^{T-1} 
\left(x_{t+1}-2x_t+x_{t-1}\right)^2,
\EEQ
where $\theta_1$ is a positive weight parameter.
For the Boolean component class, we require that
all entries of $x$ are in $\{0,\theta_2\}$, where $\theta_2$ is 
another positive parameter.
Our SD problem contains two signal class parameters, 
$\theta_1$ and $\theta_2$.
Since $\phi_3$ is not convex, the SD problem is not convex.
(Nevertheless the methods we describe below do a good job at 
approximately solving it.)

\begin{figure}
\centering
\resizebox{0.9\columnwidth}{!}{
\import{figs/}{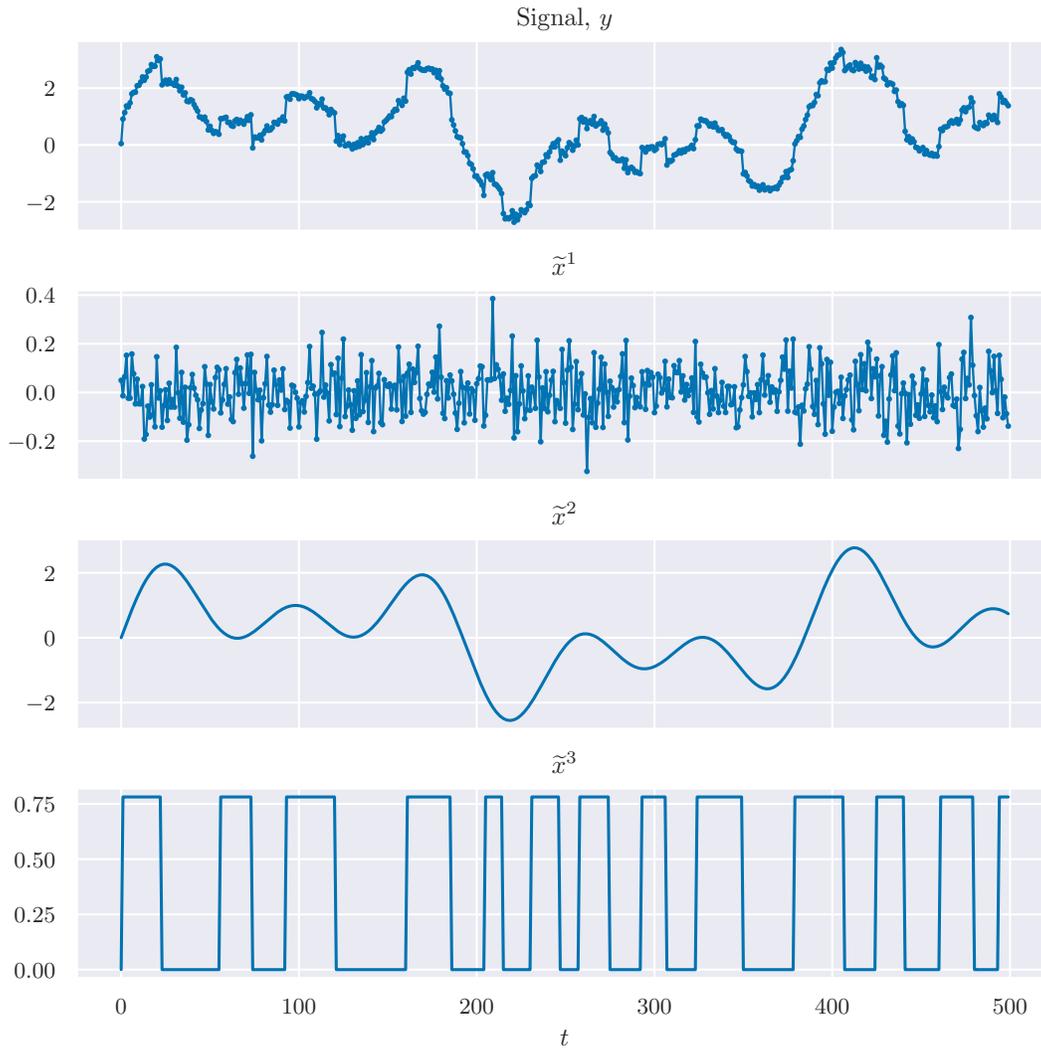}
}
\caption{Synthetic data for simple signal decomposition example.}
\label{f-example-data}
\end{figure}

\paragraph{Data generation.}
We generate a signal $y$ of length $T=500$
as a sum of three `true' signal components, one that is Gaussian noise,
one that is smooth, and one that is Boolean, \ie, takes on only
two values.
The first signal, denoted $\tilde x^1 \in \reals^T$, has IID entries
$\mathcal N(0,0.1^2)$.
The second true component is the quasiperiodic signal with three frequencies
\[
\tilde x^2_t = 
\sum_{j=1}^3 a_{j} \cos(\omega_{j} t + \delta_{j}), \quad t=1, \ldots, T,
\]
where $a_{j}>0$ are the amplitudes, $\omega_{j}>0$ are the frequencies,
and $\delta_{j} \in [0,2\pi]$ are the phases, all chosen randomly.
The last true component signal $\tilde x^3$ has the form
\[
\tilde x^3_t = \left\{ \begin{array}{ll}
\tilde \theta_2 & 
\sum_{j=1}^3 a'_{j} \cos(\omega'_{j} t + \delta'_{j}) \geq 0\\
0 &
\sum_{j=1}^3 a'_{j} \cos(\omega'_{j} t + \delta'_{j}) < 0,
\end{array}\right.
\]
for $t=1, \ldots, T$, where $a'_j$, $\omega'_j$, and $\delta'$
are a different set of amplitudes, frequencies, and phases,
also chosen randomly.
We construct the signal as
\[
y = \tilde{x}^1 + \tilde x^2 + \tilde x^3,
\]
with $\tilde \theta_2$ (the `true' value of $\theta_2$) chosen randomly.
The signal $y$ and the three true components $\tilde x^i$ are shown in 
figure~\ref{f-example-data}.
The data in this example have no missing entries.

\paragraph{Parameter search.}
We use the method described below to approximately solve the SD problem
for a grid of 21 values of $\theta_1$, logarithmically spaced between
$10^{-1}$ and $10^6$, and 21 values of $\theta_2$, linearly spaced
between $0.1$ and $2.0$, for a total of 441 different values of
the parameters.
For each of these, we evaluate the test error using 10 random selections of 
the test set as described above.  Thus all together we solved 4410 instances 
of the SD problem, which took about 13 minutes on a 2016 MacBook Pro.  
Each SD problem took about 0.17 seconds to solve. 
(We solved the problems sequentially,
but the computation is embarrassingly parallel and could have been carried
out faster using more processors.)

\begin{figure}
\centering
\resizebox{.85\columnwidth}{!}{
\import{figs/}{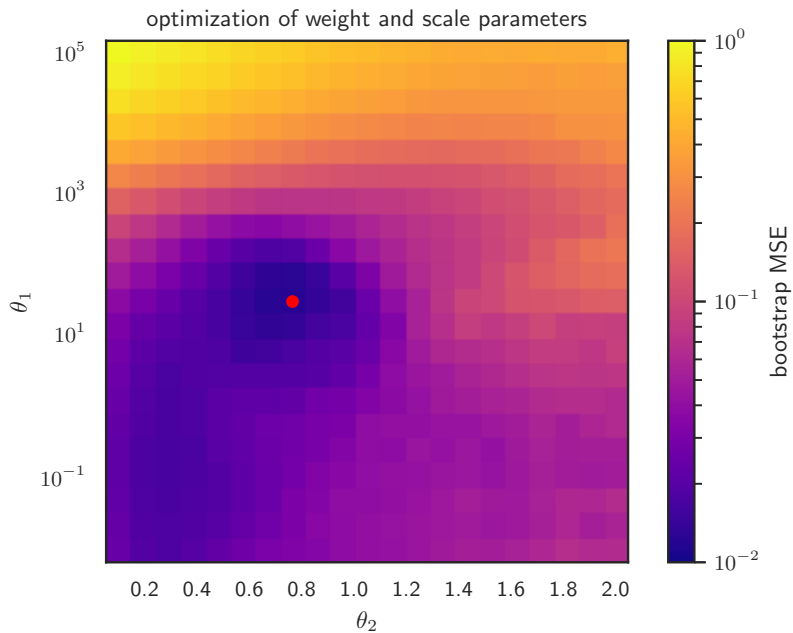}
}
\caption{Validation mean-square test error as a function of the 
parameters $\theta_1$ and $\theta_2$.}
\label{f-example-tune}
\end{figure}

The mean-square test error for the parameter grid search is shown 
as a heat map in figure~\ref{f-example-tune}. We use the final values 
$\theta_1 = 320$, $\theta_2 = 0.765$, which achieved the smallest mean-square
test error.
Having chosen $\theta_1$ and $\theta_2$, we approximately solve the SD problem 
one final time, using all the data.  

There is some discrepancy between the value we 
find $\theta_2 = 0.765$ and the true value
used to generate the data, $\tilde \theta_2 = 0.7816$,
due to the discreteness of the grid search.
(In a real application, we might do a secondary, refined grid search 
of values near the best ones found in this crude grid search.)

\begin{figure}
\centering
\resizebox{0.9\columnwidth}{!}{
\import{figs/}{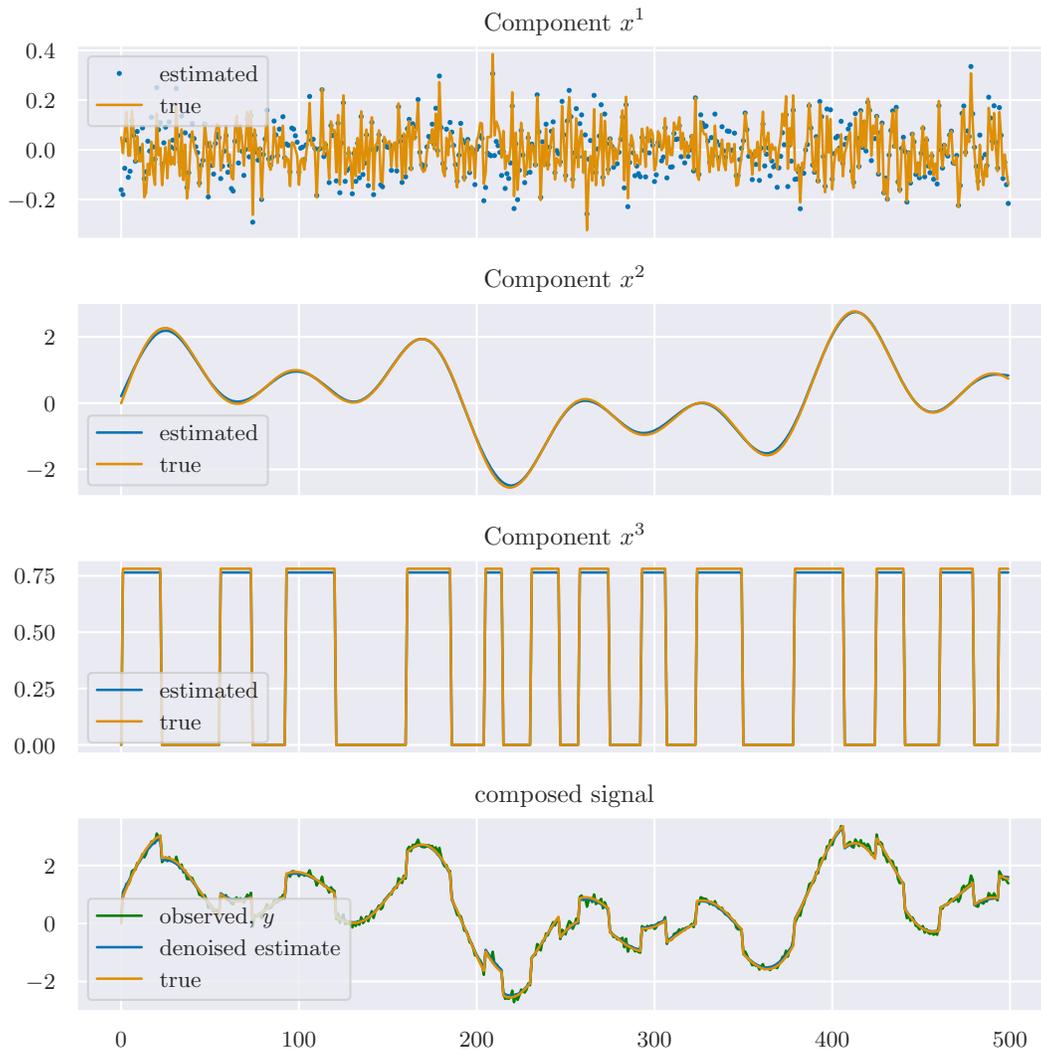}
}
\caption{Signal decomposition for simple example. The top three plots show
the true component and the esimated component.
The bottom plot shows the original signal $y$ and $x^2+x^3$, 
\ie, the decomposition without the residual component $x^1$.}
\label{f-example-decomp}
\end{figure}

\paragraph{Final decomposition.}
The final decomposition is shown in figure~\ref{f-example-decomp}. 
Evidently the decomposition is quite good.
The Boolean component is exactly reconstructed, aside from the 
slight discrepancy in its amplitude.
The smooth component is also well reconstructed,
with an RMS (root mean-square) error about 0.04.  

\section{Background and related work}\label{s-related-work}

Here we discuss a wide variety of methods that relate to the topic of signal 
decomposition, some of which are quite old. 
Many methods described below do not explicitly form an 
optimization problem, and when they do, it need not 
conform to the signal decomposition framework described in this paper.
Others methods involve minimizing a sum of loss functions for signal component 
classes subject to their sum matching an observed or given signal at known 
entries, exactly as in the proposed framework. In these cases, the discussed 
methods are often specific instances of the SD problem~\eqref{e-sd}, and these 
connections will be noted where appropriate. The SD 
formulation can be thought of as a generalization of the specific 
approaches to signal decomposition discussed in this section.

\paragraph{Regression.} Least-squares linear regression is a particular 
instance of the SD 
problem, with two component classes, a mean-square small component, 
and a component defined by a basis \eqref{e-comp-basis}, with the basis 
components
the regressors or features.
This SD problem instance admits a well-known, closed form solution~\cite[Chap.~12]
{Boyd2018}. The 
idea of solving an over-determined system of linear equations by minimizing the 
sum of the squares of the errors was proposed independently by the 
mathematicians Carl 
Friedrich Gauss and Adrien-Marie Legendre around the beginning of the 19th 
century. Statistical 
justifications for this fitting procedure were subsequently provided by Gauss, 
Laplace, Cauchy, and 
Thiele, among others~\cite{Farebrother2001}.

\paragraph{Robust regression.} Robust regression covers a variety of techniques 
to reduce model 
variance in the presence of data `outliers,' which is a term without a precise 
definition but can be 
thought of as data points that are not well explained by a linear regression 
model. Common methods include Huber 
regression~\cite{Huber1964,huber2004robust}, Theil-Sen 
estimation~\cite{Theil1950,Sen1968}, and RANSAC~\cite{Fischler1981}, which are 
included in 
the popular Python package, \texttt{scikit-learn}~\cite{scikit-learn}. In the 
SD framework, the residual component class used in linear regression is 
substituted with 
an alternative penalty function that is less sensitive to outliers. The penalty 
function formulation is 
discussed in detail in~\cite[\S 6.1 and \S 6.4]{convex_opt}.
Interestingly, the idea of minimizing the sum of absolute errors in an 
over-determined 
system of equations actually predates the development of least-squares 
minimization, 
having been proposed in the mid-18th century by 
Roger Joseph Boscovich~\cite{Fischler1981}. In the SD framework, 
robust regression is modeled using two residual classes,
one the standard mean-square small, and the other a
loss function that grows slowly for large values, like the average absolute
loss
\BEQ\label{e-avg-abs}
\phi (x) = \frac{1}{Tp} \|x\|_1 = \frac{1}{Tp} \sum_{t=1}^T \sum_{i=1}^p
|x_{t,i}|.
\EEQ
(This same loss is used as a convex heuristic for a sparse signal, \ie,
one with many entries zero.)

\paragraph{Regularized regression.} Regularized regression, 
also known as penalized regression or 
shrinkage methods, is a family of estimators that introduce an 
additional penalty term on coefficients of a linear regression problem. 
Well known examples include ridge 
regression~\cite{tihonov1963,phillips1962,hoerl1970}, lasso 
regression~\cite{Tibshirani1996}, and elastic-net regression~\cite{zou2005}. 
An overview of different regularizer functions for 
regression is given in~\cite[\S 6.3]{convex_opt}, and a review of other 
regressor 
selection methods is given in~\cite[Ch.~3--4]{Hastie2013}. In the SD framework, 
regularized regression is modeled by extending the basis class 
\eqref{e-comp-basis} with an additional loss term on the internal variable, as 
in
\[
\phi(x) = \left\{ \begin{array}{ll} \ell(z) &
x = \theta z ~\mbox{for some}~ z \in \reals^{d \times p}\\
\infty & \mbox{otherwise}.
\end{array} \right.
\]

\paragraph{Isotonic regression.}
In isotonic (or monotonic) regression we fit a given signal with a 
non-decreasing (or non-increasing) signal
~\cite{Best1990,Wu2001}.
This is a particular instance of the SD 
problem, with $p=1$ and $K=2$ component classes: a sum-of-squares small residual
and a monotone component, which has a loss function that is zero if its
argument is non-decreasing and infinite otherwise.
As efficient algorithm, with complexity linear in $T$, is included in 
\texttt{scikit-learn}~\cite{scikit-learn}. 
A detailed discussion of a linear time algorithm and the 
connection to projection operators is given in~\cite{Grotzinger1984}.

\paragraph{Trend filtering.}
Trend filtering, also called signal smoothing,
is the process of estimating a slowly varying trend 
from a scalar time series that includes rapidly varying noise.
In many cases this is also a special case of SD, 
with a mean-square residual component and 
a component that is slowly varying, for example, with a 
mean-square second difference loss function.
Trend filtering has been employed in a wide variety of applications and 
settings, 
including
astrophysics~\cite{TITTERINGTON1985}, 
geophysics~\cite{Baillie2002,Bloomfield1992,Bloomfield1992a}, social 
sciences~\cite{Levitt2004}, biology~\cite{link1994estimating}, 
medicine~\cite{Greenland1992}, image processing~\cite{Thompson1993}, 
macroeconomics~\cite{Prescott2016,Singleton1988}, and financial time series 
analysis~\cite[\S 
11]{Tsay2005}. Many specific trend filtering methods have been proposed, 
including
moving-average filtering~\cite{Osborn1995} and Hodrick-Prescott (HP) 
filtering~\cite{Prescott2016,Leser1961} being two of the most well known. 
More recently, Kim et al.\ have proposed 
$\ell_1$ trend filtering~\cite{Kim2009}, which uses as component loss function 
the $\ell_1$ norm of the second difference, which tends to result in piecewise
affine signals (see $\S$\ref{s-time-invariant-classes} for an example).
A Bayesian interpretation of trend filtering and signal 
denoising is presented in \cite{TITTERINGTON1985, 
Thompson1993, Bouman1993, Combettes2011}.

\paragraph{Seasonal-trend decomposition.}
Seasonal-trend decomposition was originally motivated by the analysis of 
economic data 
which tend to have strong seasonality; this method is arguably what most people 
think of when they 
hear the term ``time series decomposition,'' having first been proposed in the 
1920s as a natural 
extension of moving 
average smoothing~\cite{Anderson1927}. Seasonal-trend decomposition is the only 
one 
presented in the chapter on time series decomposition in Hyndman and 
Athanasopoulos ~\cite[\S 6]{Hyndman}.  A popular algorithm that implements a 
specific method
for seasonal-trend decomposition
is STL~\cite{Cleveland1990}, with packages available for Python, 
R, and Matlab~\cite{stl-python,stl-r,stl-matlab}. 

STL can be considered a specific case of the SD problem, with a scalar 
signal $y$ and $K=3$ component classes, \ie, seasonal, trend, and residual.
However, STL does not formulate the method as an 
optimization problem and uses an iterative heuristic to form the estimates of
the components. 

Modern extensions of the seasonal-trend decomposition problem have been 
introduced. In 2019, researchers from the remote sensing community proposed an 
extension that introduces a new `abrupt change' component, which is modeled as 
a piecewise linear component with a small number of 
breakpoints~\cite{Zhao2019}. Somewhat unique to this work is a focus on 
calculating uncertainty in the components and particularly the breakpoint 
locations.

\paragraph{Traditional frequency domain filtering.}
Traditional EE-style filtering (\eg, \cite{OppenheimSchafer}) can be interpreted
as a form of SD.
For example, low pass filtering decomposes a signal into a smooth component 
(the filter output)
and a small, rapidly varying component (the residual, or difference of 
the original signal and the low pass signal).
This can often be represented as SD with two components, a residual and a smooth
or low-pass component with appropriate time-invariant quadratic loss function.
A traditional filter bank can be interpreted as giving a decomposition of a 
signal into multiple components, each one 
corresponding to a different region (or sub-band) in the spectrum of the signal.

\paragraph{Sparse signal recovery.}
Sparse signal recovery is concerned with finding sparse representations of 
signals with respect to some known (typically over-complete) basis. The use of 
(convex) optimization to solve sparse signal recovery problems has a long
history with many proposed approaches, and there are some very nice 
overviews available 
in~\cite{wright2022high,McCoy2014,Tosic2011}. These methods have historically 
been applied to the problem of data compression, such as the JPEG and JPEG2000 
standards~\cite{Mallat1999,Baraniuk2010}. These methods are all 
related to the regularized linear inverse 
problem~\cite{Chandrasekaran2012},
\BEQ\label{e-bp}
\begin{array}{ll}
\mbox{minimize}   &  f(x)\\
\mbox{subject to} & Ax = y,
\end{array}
\EEQ
where the matrix $A$ and the vector $y$ are problem data, and $f$ is some 
`complexity measure' that encourages sparseness. A common variant is to relax 
the equality constraint,
\BEQ\label{e-lasso}
\begin{array}{ll}
\mbox{minimize}   &  \|Ax - y\|_2^2 + \lambda f(x).\\
\end{array}
\EEQ
When $f(x)=\|x\|_1$, \eqref{e-bp} is known as basis pursuit or 
compressed sensing, and 
\eqref{e-lasso} is the lasso, which we encountered in the 
previous paragraph. The geometry of these 
and related problems, 
specifically in the case where $f(x)=\|x\|_1$, has been extensively analyzed to 
determine when sparse signals are recoverable in~\cite{Amelunxen2014}.
The matrix $A$ generally represents the data generation process, either derived 
from known measurements or, in the case of dictionary methods, derived 
from pre-defined, parameterized waveforms, like sinusoids or wavelets.  With 
dictionary learning methods the matrix $A$ is fit to the data as 
well~\cite{Tosic2011}. When $A$ is introduced as a decision variable, 
problems~\eqref{e-bp} and~\eqref{e-lasso} are no longer convex, but there exist 
well 
established methods exists for approximately solving problems of this 
form~\cite{Udell2016}.

\paragraph{Matrix completion.}
In the basic formulation of this problem, we seek a low rank matrix $X$ 
which matches a known matrix $M$ at a set of known indices~\cite{Candes2009}. 
A closely related 
problem is (robust) principle component pursuit, in which an observed 
matrix is decomposed into a low-rank component and a sparse 
component~\cite{Candes2011,wright2022high}.

\paragraph{Convex demixing.}
Convex demixing has a long history~\cite[\S7.1]{McCoy2014a}, beginning in the 
geophysics community in the 1970s~\cite{Claerbout1973,Taylor1979}. It 
refers to the task of identifying two (or sometime more) `structured signals,' 
given only the sum of the two signals and 
information about their structures~\cite{McCoy2014,McCoy2014a}. The standard 
formulation for convex demixing is
\BEQ\label{e-cd}
\begin{array}{ll}
\mbox{minimize}   &  f(x) + \lambda g(z)\\
\mbox{subject to} & x+z=y,
\end{array}
\EEQ
where $x$ and $z$ are the decision variables, $y$ is the observed signal, 
and $\lambda$ is a regularization parameter. This is evidently a two-class, 
convex SD problem. In this 
formulation, the focus tends to be on demixing signals that are sparse in 
various senses. A classic example is the `spikes and sines problem', 
which shows up in a variety of applications including astronomy, image 
inpainting, and speech enhancement in signal 
processing~\cite{starck2010sparse,Donoho2001a}. More generally, these types of 
problems include demixing two signals that are sparse in mutually incoherent 
bases, decoding spread-spectrum transmissions in the presence of impulsive 
(\ie, sparse) errors, and removing sparse corruptions from a low-rank 
matrix. Problem \eqref{e-cd} has been deeply studied in many contexts, and much 
of the existing work has focused on finding solution methods and analyzing 
recovery bounds (\ie, uniqueness) when $f$ and $g$ are various 
sparsity-inducing matrix 
norms~\cite{Chandrasekaran2012,Bach2010}. A three-operator extension of 
\eqref{e-cd}---where one 
operator is a smooth, nonconvex function and the other two operators are convex 
functions---is studied in~\cite{Yurtsever2021}.
These are instances of the signal decomposition problem.

\paragraph{Contextually supervised source separation (CSSS).} 
This is an optimization-based framework 
for solving signal decomposition problems, in which the signal 
components are assumed to be roughly 
correlated with known basis vectors~\cite{Wytock2013}, and is very 
similar in many ways to the 
method presented in this paper. CSSS is extensible, allowing for 
different loss terms on the linear representations, component estimates, and 
linear fit coefficients. The 
SD formulation proposed in this paper is a further generalization of 
contextually supervised 
source separation, and the proposed solution method in 
\S\ref{s-solution-method} solves all instances of contextually supervised 
source separation as a subset of all SD problems.

\paragraph{Infimal convolution.} 
The infimal convolution of functions $f_i:\reals^n \to \reals$,
$i=1, \ldots, K$
denoted $f_1 \square \cdots \square f_K$, is defined as
\[
(f_1\square\cdots\square f_K)(v) = 
\inf \left\{ f_1(x^1) + \cdots + f_K(x^k) \; | \;
v = x^1 + \cdots + x^K \right\}
\]
as described (for convex functions) 
in~\cite[\S 16]{Rockafellar1970} and~\cite[\S 3.1]{Parikh2014}.
The case of nonconvex functions was considered in~\cite{Poliquin1996}. 
We see that the SD problem, with no missing data, is the problem of 
evaluating the infimal convolution of the component loss functions,
on the given signal $y$.

\paragraph{Proximal operator.}
The proximal operator of a function $f$ arises often in 
optimization, and is the basis of the solution methods described below.
The details are given below, but we note there that evaluating a 
proximal operator of the function $f$ is an SD problem (again, with no missing 
data) with a 
mean-square loss and the loss $f$.

\paragraph{Our contribution.}
We present a common formulation for describing generalized signal decomposition 
problems as optimization problems. This treatment fully embraces the handling 
of missing data and is extensible to many new problem formulations. When no 
data is missing, this framework exactly represents 
many methods described in this section as specific cases. Aside from the use of 
a masked proximal operator (described below), the proposed 
solution method is based on well known algorithms, block coordinate descent 
(BCD) and 
the alternating direction method of multipliers (ADMM). We note that ADMM is a 
common choice for convex demixing problems~\cite{McCoy2014}, and that we are 
able to apply 
BCD to these problems because of the structure that we enforce on the signal 
decomposition models that the first term be a mean-square-small residual term.

\section{Solution methods} \label{s-solution-method}

In this section we describe two related methods for solving the 
SD problem (when it is convex), and approximately solving it 
(when it is not convex).  Both rely on the masked proximal operators
of the component class losses, but aside from that, they are small variations 
of block coordinate descent and the alternating direction method of 
multipliers. Finally, we describe a hybrid algorithm, combining the BCD and 
ADMM approaches.

\subsection{Masked proximal operator }\label{s-prox}
Recall that the \emph{proximal operator} \cite{Moreau1962,Parikh2014} of 
$\phi_k$
is defined as
\BEAS
\prox_{\phi_k}(v) &=&
\argmin_x \left( \phi_k(x) + \frac{\rho}{2}\|x-v\|_F^2  \right) \\
&=& \argmin_x \left( \phi_k(x) + \frac{\rho}{2} \sum_{t,i} (x_{t,i}-v_{t,i})^2 
\right),
\EEAS
where $\rho$ is a positive parameter, and $v \in \reals^{T \times p}$.
When $\phi_k$ is convex, 
the function minimized is strictly convex, so there is a unique argmin.
When $\phi_k$ is not convex, 
there can be multiple argmins; we simply choose one.

The \emph{masked proximal operator} is defined as
\BEAS
\mprox_{\phi_k}(v) &=& \argmin_x \left( \phi_k(x) + \frac{\rho}{2}
\|\mask(x-v)\|_2^2  \right)\\
&=& \argmin_x \left( \phi_k(x) + \frac{\rho}{2} \sum_{(t,i)\in \mathcal K}
(x_{t,i}-v_{t,i})^2 \right).
\EEAS
Roughly speaking, it is the proximal operator, with the norm term
only taken over known entries.  (The masked proximal operator 
depends on $\mathcal K$, but we suppress this dependency to keep the notation
lighter.)  
The function minimized in the masked proximal operator need not have
a unique minimizer, even when $\phi_k$ is convex.  In this case, we simply 
pick one.

When the function $\phi_k$ takes on the value $\infty$ 
(\ie, encodes constraints),
the point $x=\mprox_{\phi_k}(v)$ is feasible, \ie, 
satisfies $\phi_k(x)< \infty$.
We also note that $\mprox_{\phi_k}(v)$ does not depend 
on $v_{t,i}$ for $(t,i) \in \mathcal U$, so we have 
\BEQ\label{e-mprox-MM}
\mprox_{\phi_k}(v) = \mprox_{\phi_k}(\mask^* \mask v).
\EEQ

When there are no unknown entries, \ie, 
$\mathcal U= \emptyset$,
the masked proximal operator reduces to the standard proximal operator.
There is another simple connection between the proximal operator and the masked
proximal operator.
Starting with a loss function $\phi$, we define the function
\[
\tilde \phi(z) = \inf \{ \phi(\mask^* \mask z + u) \mid \mask u =0 \},
\]
which is, roughly speaking, 
the original loss function where we minimize over 
the unknown entries in $y$.  If $\phi$ is convex, so is $\tilde \phi$, since
it is its partial minimization \cite[\S 3.2.5]{convex_opt}.  
The masked proximal operator is then
\[
\mprox_\phi (v) = \prox_{\tilde \phi}(v),
\]
the proximal operator of the partially minimized loss function.

For many component loss functions we can work out the masked proximal operator
analytically.
In many other cases we can compute it with reasonable cost, often 
linear in $T$, the length of the signals.
The monographs \cite[\S 6]{Parikh2014} and \cite{Boyd2011} discuss the 
calculation 
of proximal operators in depth and list many well known results. 
Many closed form proximal operators are listed in the 
appendix of \cite{Combettes2011}.
Many of these have straightforward extensions to the masked proximal operator.

As a final generalization, we introduce the weighted proximal 
operator, which we define as
\[\wprox_{\phi_k}(v) = \argmin_x \left( \phi_k(x) + \frac{\rho}{2} 
\sum_{(t,i)\in \mathcal K} w_{t,i}
(x_{t,i}-v_{t,i})^2 \right),\]
with nonnegative weights $w_{t,i}\in\reals_{+}$ for all $(t,i)\in\mathcal K$. 
The weighted proximal operator arises in the evaluation of certain masked 
proximal 
operators, as discussed in~\S\ref{s-convex-quad} and~\S\ref{s-common-term}.
When all the weights are one, the weighted proximal operator coincides with
the masked proximal operator.

\paragraph{Proximal operator as SD problem.}
We note that the proximal operator itself can be seen as a simple instance
of an SD problem, with $v$ playing the role of $y$, and components $x$ and 
$v-x$,
with associated loss functions $\phi_k$ and $(\rho/2)\|\cdot\|_F^2$,
respectively.
The masked proximal operator is the version of this signal decomposition
problem with missing entries in $v$.

Thus, evaluating the masked proximal operator is the same as solving a simple
SD problem with two components, one of which is scaled mean-square small.
Our algorithms, described below, solve (or approximately solve) the general SD
problem by iteratively solving these simple two component SD problems for 
each component.

\paragraph{Surrogate gradient.}
When $\phi$ is convex, the optimality condition for 
evaluating the masked proximal operator $x=\mprox_\phi(v)$ tells us that
\BEQ\label{e-prox-grad}
g = \rho \mask^* \mask (v - x) \in \partial \phi(x),
\EEQ
where $\partial \phi(x)$ is the subdifferential (set of all subgradients)
of $\phi$ at $x$.
So evaluating the masked proximal operator at a point $v$ 
automatically gives us a 
subgradient of the loss at the image point $x=\mprox_\phi(v)$.
When $\phi$ is not convex, we can interpret $g$ in \eqref{e-prox-grad} 
as a surrogate gradient.

\paragraph{Stopping criterion.}
In both algorithms, $x^2, \ldots, x^K$ are
found by evaluating the loss function masked proximal operators, \ie,
\[
x^k = \mprox_{\phi_k} (v^k), \quad k=2, \ldots, K,
\]
for some $v^k$.
(The particular $v^k$ used to find $x^k$ depend on which algorithm is used,
but each of them satisfies $v^k = \mask^*\mask v^k$, \ie, they are zero
in the unknown entries of $y$.)
We define $x^1 = \mask^*\mask(y-x^2- \cdots - x^K)$, so
$x^1 , \ldots , x^K$ are feasible and $x^1 = \mask^*\mask x^1$.

We combine \eqref{e-dual-feas-subg} with \eqref{e-prox-grad} and define
the optimality residual $r$ as
\BEQ\label{e-opt-resid}
r = \left( \frac{1}{K-1} \sum_{k=2}^K \left\| 
\rho\mask^*\mask(v^k-x^k) - \frac{2}{Tp}x^1 \right \|_F^2 \right)^{1/2},
\EEQ
which can be written as
\[
r = \left( \frac{1}{K-1} \sum_{k=2}^K \left\| \mask\left(
\rho(v^k-x^k) - \frac{2}{Tp} x^1 \right) \right \|_2^2 \right)^{1/2}.
\]
When $r=0$ and the losses are convex, $x^1, \ldots, x^K$ are optimal.

Both algorithms use the standard stopping criterion
\BEQ\label{e-stopping-crit}
r \leq \epsilon^\text{abs} + \epsilon^\text{rel} \left\| \frac{2}{Tp}x^1 
\right\|_F =
\epsilon^\text{abs} + \epsilon^\text{rel} \left\| \frac{2}{Tp} \mask x^1 
\right\|_2,
\EEQ
where $\epsilon^\text{abs}$ and
$\epsilon^\text{rel}$ are specified positive absolute and relative 
tolerances.

\subsection{Block coordinate descent algorithm}

The BCD algorithm repeatedly minimizes the objective in \eqref{e-sd-no-x1},
\[
\frac{1}{Tp} \left\| \mask y- \mask x^2 - \cdots - \mask x^K
\right\|_2^2 + \phi_2(x^2) + \cdots + \phi_K(x^K),
\]
over a single (matrix) variable $x^k$, holding the other variables fixed.
Minimizing the objective over $x^k$, with $x^i$ fixed for $i \neq k$,
is the same as evaluating the masked proximal operator of $\phi_k$:
\[
x^k = \mprox_{\phi_k} \left( y - \sum_{i \neq k} x^i \right)
\]
with parameter $\rho = 2/(Tp)$.
(Note that the masked proximal operator does not depend on the entries of 
its argument that are unknown in $y$.)
There are many choices for the sequence in which we minimize over the 
variables, but we will use the simplest round-robin method,
updating $x^2$, then $x^3$, and on to $x^K$, and then back to $x^2$ again.
This gives the SD-BCD algorithm described below,
with superscript $j$ on the variables denoting iteration number,
where an iteration consists of one cycle of 
(successively) minimizing over $x^2, \ldots, x^K$.

\begin{algdesc}{\sc Block coordinate descent algorithm 
for SD problem (SD-BCD)}\label{alg-bcd}

\emph{Initialize.} Set $(x^k)^0$, $k=2,\ldots, K$, as some initial estimates.

\textbf{for} iteration $j=0,1,\ldots$

\textbf{for} component class $k=2,\ldots, K$

\emph{Update a component using masked proximal operator.}
\[
(x^k)^{j+1} = \mprox_{\phi_k} \left(y - \sum_{i < k} (x^i)^{j+1} -
\sum_{j>k} (x^i)^j \right).
\]
\end{algdesc}

In SD-BCD we use the most recently updated value for the other 
components, in Gauss-Seidel fashion.
Since we fix $\rho = 2/(Tp)$, this algorithm contains no parameters to tune.
Note that SD-BCD accesses the component class loss 
functions only through their masked proximal operators; 
in particular we never evaluate $\phi^k$ or its derivatives.

\paragraph{Stopping criterion.}
We evaluate the stopping criterion \eqref{e-stopping-crit} at the 
end of each iteration, using $x^1= \mask^*\mask(y-x^2- \cdots - x^K)$
and $v^k$ the argument of the proximal operator in SD-BCD.

\paragraph{Convergence.}
SD-BCD is evidently a descent algorithm, \ie, the objective is nonincreasing in 
each iteration. 
(In fact, it is nonincreasing after each update of one of the components.)
Well known simple examples show that block coordinate descent need not converge
to an optimal point even when the objective is convex. There is a large 
body of literature on the convergence of block coordinate descent type methods. 
Some recent review papers 
inlcude~\cite{Wright2015,Beck2013,Richtarik2014} and a classic textbook 
that addresses the topic is~\cite[\S3.7]{Bertsekas2016}. 
These convergence proofs often rely on randomly permutating the block
update order, but we have
found this has no practical effect on the convergence of SD-BCD.
None of cited literature exactly proves the convergence 
of the algorithm presented here, so 
we give a simple proof that any fixed 
point of SD-BCD must be optimal, when the losses are all convex.
When one or more loss functions are not convex, the algorithm may (and often 
does)
converge to a non-optimal stationary point.

\paragraph{Fixed point of SD-BCD.}
Here we show that if $x^2,\ldots, x^K$ are a fixed point
of SD-BCD, and the losses are all convex, then the decomposition is 
optimal.
If these variables are a fixed point, then for $k=2, \ldots, K$,
\[
x^k = \mprox_{\phi_k} \left( y-\sum_{i \geq 2, ~i \neq k} x^i \right).
\]
From these and \eqref{e-prox-grad} we find that for $k=2, \ldots, K$,
\BEAS
g^k &=& \frac{2}{Tp} \mask^*\mask
\left(y - \sum_{i \geq 2, ~ i \neq k} x^i -x^k \right)\\
&=&
\frac{2}{Tp}  \mask^*\mask
\left(y- \sum_{i=2}^K x^i\right)\\
&=& 
\frac{2}{Tp}x^1 \\
&\in& \partial \phi_k(x^k),
\EEAS
where in the third line we use $x^1 = \mask^*\mask(y-\sum_{i=2}^K x^i)$.
This is the optimality condition \eqref{e-dual-feas-subg}.

\subsection{ADMM algorithm}
Here we introduce an operator splitting method for the SD problem.
The particular operator splitting method we use is the
alternating directions method of multipliers (ADMM) 
\cite{Glowinski1975,Gabay1976,Boyd2011}.
The ADMM algorithm we develop for the SD problem is closely related
to the sharing problem~\cite[\S 7.3]{Boyd2011} and
the optimal exchange problem~\cite[\S 7.3.2]{Boyd2011}, but not the same.
The algorithm uses a scaled dual variable $u \in \reals^q$,
and we denote iteration number with the superscript $j$.

\begin{algdesc}{\sc ADMM for SD problem (SD-ADMM)}\label{alg-admm}

\emph{Initialize.} Set $u^0 = 0\in \reals^q$,
and $(x^k)^0 \in \reals^{T \times p}$, $k=1, \ldots, K$,
as some initial estimates

\textbf{for} iteration $j=0,1,\ldots$
\begin{enumerate}
\item \emph{Evaluate masked proximal operators of component classes in 
parallel.}
\[
(x^k)^{j+1} = \mprox_{\phi_k}((x^k)^j - 2 \mask^* u^j), \quad k=1, \ldots, K.
\]
\item \emph{Dual update.}
\[
u^{j+1} = u^j + \frac{1}{K}
\left(\sum_{k=1}^K \mask(x^k)^{j+1} - \mask y\right).
\]
\end{enumerate}
\end{algdesc}

A detailed derivation of this algorithm is given in appendix 
\S\ref{s-sd-admm-deriv}.
Unlike BCD, SD-ADMM is not a descent method.  It is also not a feasible 
method: the iterates satisfy the consistency constraint
$\mask y = \mask x^1 + \cdots + \mask x^K$ only in the limit.

\paragraph{Interpretations.}
From the dual update, we see that $u^j$ is the running sum of
the residual in the consistency constraint, scaled by $1/K$; 
this term is used in the argument of the masked proximal operator to
drive $x^k$ to optimality.

\paragraph{Convergence with convex losses.}
When $\phi_k$ are all convex,
SD-ADMM converges to a solution, and $\rho u^j$ converges to
an optimal dual variable $\nu$ \cite[\S 3.2]{Boyd2011}.
In particular, the consistency constraint \eqref{e-decomp} holds asymptotically.

\paragraph{Convergence with nonconvex losses.}
When any of the loss functions is nonconvex, there are no convergence 
guarantees at all.  The ADMM algorithm need not 
converge, and if it converges it need not converge to a solution of the SD
problem.  
But it has been observed in practice that ADMM, when applied to nonconvex 
problems, often converges to a useful value,
which in this case is a useful signal decomposition; 
see, \eg, ~\cite[\S 9]{Boyd2011}.

\paragraph{Stopping criterion and final decomposition.}
The consistency constraint generally does not hold for the iterates.
To obtain a decomposition that satisfies the consistency constraint,
we can simply absorb the residual in the consistency constraint into $x^1$ to
obtain a feasible signal decomposition.
We can then evaluate the residual in \eqref{e-stopping-crit}, with
$v^k$ the arguments of the proximal operators in step~1 of SD-ADMM.

\paragraph{Choice of $\rho$.}
When the problem is convex, SD-ADMM
converges to a solution for any positive value of 
the algorithm parameter $\rho$, although the practical convergence speed 
can be affected the choice of $\rho$.
The natural value $\rho=2/(Tp)$ seems to give good performance in 
practice.
When the problem is not convex, the choice of $\rho$ is more critical,
and can affect whether or not the algorithm converges, and when it converges,
the decomposition found.
For such problems too, the natural choice $\rho = 2/(Tp)$ seems to often
give good results, although we have found that scaling this value can 
improve the practical convergence for some nonconvex problems. 
We take $\rho = 2 \eta/(Tp)$, with $\eta$ in the range between 0.5 and 2.

\subsection{Hybrid algorithms}
\paragraph{Comparison of SD-BCD and SD-ADMM.}
For convex SD problems, SD-BCD often outperforms SD-ADMM, but not by much.
For nonconvex SD problems, we have found that SD-ADMM often
outperforms SD-BCD in the quality of the decomposition found.
Specifically, CD-BCD often ends up converging to a poor local minimum, 
whereas SD-ADMM
is able to find a much better (lower objective) decomposition.
On the other hand, for nonconvex SD problems,
one or two iterations of SD-BCD, starting from the decomposition found 
by SD-ADMM, can lead to a modest improvement
in the objective value found. (These iterations cannot increase 
the objective, since SD-BCD is a descent method.)

\paragraph{Hybrid methods.}
A reasonable strategy, and the default in our implementation,
is to use SD-BCD if the SD problem is convex.
If the SD problem is nonconvex, the default uses SD-ADMM (with scale 
factor $\eta = 0.7$) until convergence,
and then follows this with SD-BCD, again run until convergence
(quite often, but not always, only a few iterations).
This hybrid method seems to work well on a wide variety of SD problems. 

\begin{figure}
\centering
\resizebox{0.95\columnwidth}{!}{
\import{figs/}{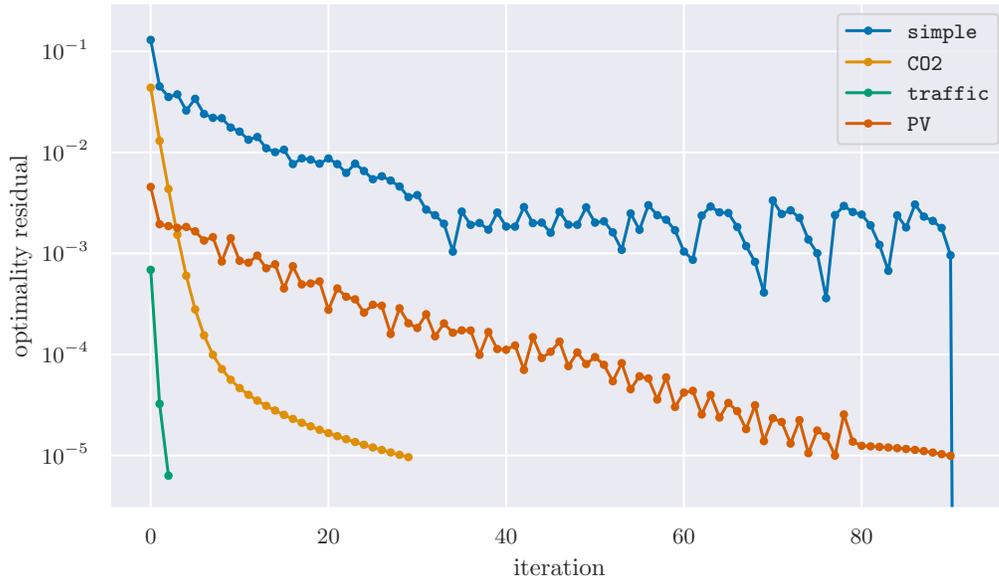}
}
\caption{Residual versus iteration number for the 4 
numerical examples given in \S\ref{s-simple-example}, \S\ref{s-CO2}, 
\S\ref{s-traffic}, and \S\ref{s-pv-fleet} respectively.}
\label{f-convergence}
\end{figure}

\paragraph{Numerical examples.} 
In this paper we consider four numerical examples, summarized in
table~\ref{t-examples}. 
They include convex and nonconvex problems, and range from small to 
large, with the SD problem in \texttt{PV} having over 700,000 variables.
We use these examples to illustrate the convergence of the hybrid algorithm.
In figure~\ref{f-convergence} we plot the residual
\eqref{e-opt-resid} versus iteration number for these four
problems. 

We see rapid and monotonic convergence for problems \texttt{CO2} and 
\texttt{traffic},
which are convex.
For \texttt{simple} and \texttt{PV}, which are nonconvex, we can see 
the switch to SD-BCD at the end, with a sharp reduction in residual 
in \texttt{simple} in just a few iterations, and a smoother 
reduction of residual over 12 iterations in \texttt{PV}.
None of the examples requires more than 100 iterations to converge. 

\begin{table}
\centering
\caption{Summary of numerical examples}
\begin{tabular}{l|c|c|r|c|r|r|c}
Name & Section & $K$ & $T$ & $p$ & 
Size ($KTp$) & $q$ & Convex \\
\hline
\texttt{simple} & \S\ref{s-simple-example} & 3 & 500 & 1 & 1,500 & 500 & no \\
\texttt{CO2} & \S\ref{s-CO2} & 3 & 2,459 & 1 & 7,377 & 2,441 & yes \\
\texttt{traffic} & \S\ref{s-traffic} & 5 & 105,552 & 1 & 527,760 & 101,761 & 
yes \\
\texttt{PV} & \S\ref{s-pv-fleet} & 5 & 20,212 & 7 & 707,420 & 135,899 & no \\
\end{tabular}
\label{t-examples}
\end{table}

\section{Component class attributes} \label{s-class-attr}

In this section we describe some very basic attributes that component class
losses can have.

\subsection{Separability}
A component class loss function can be separable across time, or entries, or 
both.

\paragraph{Time-separable losses.}
A function $\phi: \reals^{T \times p}\to \reals \cup \{\infty\}$ is 
separable across time if it has the form 
\[
\phi(x) = \sum_{t=1}^T \ell_t(x_t)
\]
for some functions $\ell_t:\reals^p \to \reals \cup \{\infty\}$, $t=1, \ldots, 
T$.
It is common for the loss functions to not depend on $t$, 
in which case we say it is time-invariant.
A simple example is the mean-square loss \eqref{e-ms-small},
with $\ell_t(x_t) = \frac{1}{Tp}\|x_t\|_2^2$ for all $t$.

\paragraph{Entry-separable losses.}
A component class function $\phi$ is separable across entries if it has the form
\[
\phi(x) = \sum_{i=1}^p \ell_i(\breve x_i)
\]
for some functions $\ell_i:\reals^T \to \reals \cup \{\infty\}$, $i=1, \ldots, 
p$,
where $\breve x_i$ is the $i$th column
of $x$ (which can be interpreted as a scalar time series), the $i$th entry of 
the vector
time series $\{x_t\}$.
Here too it is common for the loss function to not depend on $i$, in which case
we say it is symmetric (in the entries of $x_t$).
The mean-square loss \eqref{e-ms-small} is symmetric (in addition to being 
time-separable).

\paragraph{Separability and proximal operators.}
Separability reduces the complexity of evaluating the masked proximal 
operator.  
For example if $\phi$ is separable across time, say, $\phi(x) = \sum_{t} 
\ell_t(x_t)$,
its masked proximal operator is 
\[
\mprox_\phi (v) = \left[ \begin{array}{c} 
\mprox_{\ell_1} (v_1)^T \\
\vdots \\
\mprox_{\ell_T} (v_T)^T
\end{array}\right],
\]
\ie,
we can evaluate the masked proximal operator in parallel for each 
time $t=1, \ldots, T$.  (Note the masked proximal operator for $t$ depends on
the missing data for that time period.)

\subsection{Time-invariance}
Time-invariance or shift-invariance is another important attribute.
We let $M<T$ denote the memory of the loss function $\phi$.
We say $\phi$ is time-invariant if it has the form
\[
\phi(x) = \sum_{t=1}^{T-M+1} \ell (x_{t:t+M-1}),
\]
where $x_{t:t+M-1}$ is the $M \times p$ slice of $x$, that includes rows $t, 
\ldots, t+M-1$,
and $\ell : \reals^{M \times p} \to \reals \cup\{\infty\}$ is the slice loss.
Thus, a time-invariant component class loss is sum of the slice loss, applied 
to all $M$-long
slices of its argument.
With this definition, a time-separable time-invariant loss
is a special case of time-invariance, with memory $M=1$.

The second-order mean-square smooth loss \eqref{e-ss-second} is a simple 
example of a 
time-invariant component class loss, with $M=3$.
As another example, consider the class of $P$-periodic signals, with loss
\BEQ\label{e-P-periodic}
\phi(x) = \left\{ \begin{array}{ll}
0 & x_{t+P}=x_t, \quad t=1, \ldots, T-P,\\
\infty & \mbox{otherwise},
\end{array}\right.
\EEQ
which has memory $M=P+1$.

\subsection{Convex quadratic}\label{s-convex-quad}
A loss is convex quadratic if it has the form
\BEQ\label{e-quadratic}
\phi(x) = \left\{ \begin{array}{ll}
(1/2) x_:^T P x_: + q^T x_: + r & A x_: = b\\
\infty & \mbox{otherwise},
\end{array}\right.
\EEQ
where $x_:\in \reals^ {Tp}$ is a vector representation of $x$ (and $ x_:^T$ is 
its transpose),
$P \in \reals^{Tp \times Tp}$ is symmetric positive semidefinite,
$q \in \reals^{Tp}$, $r \in \reals$, $A \in \reals^{L \times Tp}$,
and $b\in \reals^L$. Thus $\phi$ is convex quadratic, with some 
equality constraints.
We have already encountered a few examples of convex quadratic loss functions,
such as mean-square small and mean-square smooth.

As a more interesting example, consider the $P$-periodic smooth loss,
defined as
\BEQ\label{e-P-periodic-smooth}
\phi(x) = 
 \frac{1}{Pp} \left( \|x_2-x_1\|_2^2 + \cdots + \|x_P-x_{P-1}\|_2^2 + 
\| x_1 - x_P \|_2^2 \right),
\EEQ
provided $x$ is $P$-periodic, \ie,
$x_{t+P}=x_t$ for $t=1, \ldots, T-P$, and $\phi(x) = \infty$, otherwise.
This is the same as the $P$-periodic loss \eqref{e-P-periodic},
with mean-square smoothness, taken circularly.

\paragraph{Masked proximal operator of convex quadratic loss.}
The masked proximal operator of a convex quadratic loss function 
can be efficiently evaluated; more precisely, after the first
evaluation, subsequent evaluations can be carried out more efficiently.
Evaluating the masked proximal operator involves minimizing a convex quadratic
function subject to equality constraints, which in turn can be done by solving 
a set
of linear equations, the KKT (Karush-Kuhn-Tucker) equations 
\cite[\S 16]{Boyd2018}.
If we cache the factorization used to solve this set of linear equations
(\eg, the $LDL^T$ factorization of the coefficient matrix), subsequent 
evaluations
require only the so-called back-solve step, and not the factorization.
This idea is often exploited in ADMM; see \cite[\S 4.2]{Boyd2011}.

\paragraph{Weighted proximal operator.} In evaluating the masked proximal 
operators of certain convex quadratic loss functions, it can more 
computationally efficient to evaluate a related weighted proximal operator. 
This is seen commonly with loss functions that are $P$-periodic. In this case, 
the solution to the masked proximal operator may be found by evaluating a 
smaller weighted proximal operator. Specifically, the weighted proximal 
operator is evaluated over a vector $z\in\reals^{P\times p}$, representing a 
single period of component. The input to this smaller proximal operator is the 
original input, averaged across periods, using only the available data,~\eg, 
the entries in $\mathcal K$. The weights are defined as the number of real 
entries used in each averaging operation, divided by the total possible number 
of entries. (Some additional care must be taken here when evaluating signals 
that are not an even multiple of the period length.)

\subsection{Common term}\label{s-common-term}
Another common attribute of a component class is when it represents a common
term across the entries of the signal.  The loss has the form
\BEQ\label{e-common-term}
\phi(x) = \left\{ \begin{array}{ll}  \tilde \phi(z) & x_t = z_t \ones, \quad 
t=1,\ldots, T\\
\infty & \mbox{otherwise},
\end{array} \right.
\EEQ
where $\tilde \phi: \reals^T \to \reals \cup \{\infty\}$ is a loss function 
for a scalar signal.
Roughly speaking, this component class requires that all entries of 
$x$ (\ie, its columns) are the same, and uses a scalar-valued signal loss 
function on
the common column.
If $\tilde \phi$ is separable, then $\phi$ is separable across time.

The proximal operator of such a $\phi$ is readily found in terms of 
the proximal operator of $\tilde \phi$.  It is
\[
\prox_\phi (v) = \prox_{\tilde \phi} ((1/n) v\ones)\ones^T.
\]
In words: to evaluate the proximal operator for a common term 
loss function, we first average the columns of $v$, then apply 
the proximal operator of $\tilde \phi$, and finally 
broadcast the result to all columns. 

The masked proximal operator is a bit more complex. Each row can have a 
different number of entries in the known set, so 
the average across columns must be taken with respect to the number of real 
entries in the row instead of the number of columns. However, to make use of 
the scalar formulation $\tilde \phi(z)$, we must invoke the weighted proximal 
operator,
\BEAS
\mprox_{\phi}(v) &=&
\argmin_x \left( \tilde \phi(z) + \frac{\rho}{2}\sum_{t,i\in\mathcal K}(x_{t,i} 
- v_{t,i})^2  \right),\quad\mbox{s.t. }x_t = z_t\ones \\
&=& \wprox_{\tilde \phi}(\mathbf{ravg}(v))\ones^T,
\EEAS
where $\mathbf{ravg}:(\reals\cup\{?\})^{T\times 
p}\rightarrow(\reals\cup\{?\})^T$ is the row-wise 
average of the matrix $v$, over only the 
known entries. (If a row has no known entries, the function returns 
$?$ for that time index.) The weights are the number of known entries 
used in each averaging operation, divided by the total possible number of 
entries.

\section{Component class examples} \label{s-classes}

There is a wide variety of useful component classes; in this section we 
describe some typical examples.
In most cases the proximal operator of the loss is well known,
and we do not give it;
we refer the reader to other resources, 
such as \cite{Combettes2011,Parikh2014,Boyd2011}.
When the loss function is convex, but an analytical method to evaluate
the proximal operator is not known, we can always fall back on a 
numerical method, \eg, using CVXPY \cite{diamond2016cvxpy,agrawal2018cvxpy}.
In a few cases where we believe our method of evaluating the proximal 
operator is new, we give a short description of the method.

\subsection{Time-separable classes}

Time-separable classes are given by the loss functions $\ell_t$ on $\reals^p$.
We have already seen the mean-square small class, with loss $\ell_t(u) = 
\frac{1}{Tp}\|u\|_2^2$,
and the finite set class, which requires that $x_t$ be one of a given set of 
values.
We mention a few other examples in this section.

\paragraph{Value constraint component classes.}
As an extension of the finite value class, we require that $x_t \in \mathcal 
S_t$, where
$\mathcal S_t \subset \reals^p$ is some given set.  
If $\mathcal S_t$ are all convex, we have a convex
loss function.  Simple convex examples include the nonnegative component class, 
with $\mathcal S = \reals_+^p$, and the vector interval signal class, with 
$\mathcal S = \{ u\mid x_t^\text{min} \leq u \leq x_t^\text{max} \}$, where 
the inequality is elementwise and
$x_t^\text{min}$ and $x_t^\text{max}$ are given lower and upper limits on the 
entries of the signal (which can be parameters).
In addition to the constraint $x_t \in \mathcal S_t$,
we can add a nonzero penalty function of $x_t$ to the objective. 

\paragraph{Mean-square close entries.}
The loss 
\BEQ\label{e-entries-close}
\ell(u) = \frac{1}{p} \sum_{i=1}^p (u_i - \mu)^2, \qquad
\mu  = \frac{1}{p} \sum_{i=1}^p u_i,
\EEQ
which is the variance of the entries of the vector $u$, defines the 
mean-square close entries class.
If we scale this class by a very large weight, this gives
an approximation of the common term class \eqref{e-common-term} 
(with $\tilde \phi=0$), in which the 
entries of the signal must be the same for each $t$.

\paragraph{Robust losses.}
We can modify the sum of squares loss so the component class can include 
signals with 
occasional outliers, using so-called robust losses, which grow linearly for 
large 
arguments, when they are convex, or sub-linearly when they are not.
One well-known examples is the Huber loss, defined as 
\[
\ell(u) = \sum_{i=1}^p H(u_i), \qquad
H(a) = \left\{ \begin{array}{ll} a^2 & |a| \leq M \\
M(2|a|-M) &  |a| > M,
\end{array}\right.
\]
where $M>0$ is a parameter \cite[\S 6.1.2]{convex_opt}.
An example of a nonconvex robust loss is the log Huber loss,
\[
\ell(u) = \sum_{i=1}^p \tilde H(u_i), \qquad
\tilde H(a) = \left\{ \begin{array}{ll} a^2 & |a| \leq M \\
M^2(1 + 2 \log(|a|/M) ) &  |a| > M.
\end{array}\right.
\]

\paragraph{Convex sparsity inducing losses.} 
The loss function $\ell(u) = \|u\|_2$ (note that this norm is not squared) 
leads to
vector-sparse (also called block sparse) component signals, \ie, ones for which 
for many 
values of $t$, we have $x_t=0$.
In machine learning this is referred to as group lasso \cite[\S 
3.8.4]{Hastie2013}.
With this loss, we typically find that when $x_t \neq 0$, all its entries are
nonzero.
The loss function $\ell(u)  =\|u\|_1$, sum-absolute small
component class, tends to yield signals that are 
component-wise sparse, \ie, for many values of $(t,i)$, we 
have $x_{t,i}=0$.

\paragraph{Non-convex sparsity inducing losses.} 
The most obvious one is the cardinality or number of nonzeros loss,
with $\ell(u)$ being the number of nonzero entries in $u$ (or, in the vector 
version,
$0$ if $u=0$ and $1$ otherwise).
In this case the overall loss $\phi(x)$ is the number of nonzero values of 
$x_{t,i}$.
A variation is to limit the number of nonzeros to some given number, say, $r$, 
which gives
the $r$-sparse signal component class.

These losses are nonconvex, but have well-known analytic expressions for their
proximal operators.
For example when the loss is the number of nonzero entries in $x$,
the proximal operator is so-called hard thresholding \cite[\S 9.1.1]{Boyd2011},
\[
\prox_{\phi_k} (v)_{t,i} = \left\{ \begin{array}{ll} 
0 & |v_{t,i}| \leq \sqrt{2/\rho}\\
v_{t,i} & |v_{t,i}| > \sqrt{2/\rho},
\end{array}\right. \quad t=1, \ldots, T, \quad i=1, \ldots, p.
\]

\paragraph{Quantile small.} 
The quantile loss~\cite{Koenker1978,Koenker2001} is a variation on the $\ell_1$ 
loss $\|u\|_1$,
that allows positive and negative values to be treated differently:
\BEQ\label{e-quantile}
\ell(u) = \sum_{i=1}^p \left( |u_{i}|+(2\tau - 1) u_{i}\right),
\EEQ
where $\tau\in(0,1)$ is a parameter. 
For $\tau=0.5$, this class simplifies to sum-absolute small. 
(Its proximal operator is given in \cite[\S2.2,\S 6.5.2]{Parikh2014}.)


\subsection{Time-invariant classes}\label{s-time-invariant-classes}

Any time separable loss for which $\ell_t$ do not depend on $t$ is 
time-invariant.  We give a few other examples here.

\paragraph{Index-dependent offset.}
In the common term class \eqref{e-common-term},
the entries of signals are the same.
The index-dependent offset class is analogous: Its signals are 
different for different indexes, but the same over time.
It is given by $\phi(x) = 0$ if for some $z$, $x_t = z$ for all $t$,
where $z \in \reals^p$, and $\infty$ otherwise.  Of course we can add 
a penalty on $z$.
This loss is time-invariant, with a memory of one.

\paragraph{Higher order mean-square smooth component classes.}
We have already mentioned the mean-square smooth class
which uses the first-order difference (\ref{e-ms-smooth}), and its
extension to the second-order difference (\ref{e-ss-second}).  
Higher order mean-square smooth classes use higher order differences.

%

\paragraph{Mean-absolute smooth.}
Replacing the mean-square penalty in mean-square first-order smooth classes
with a average-absolute penalty yields a components whose signal entries
are typically piecewise constant.  With the second-order difference,
\BEQ\label{e-l1-trend}
\phi(x) = 
 \frac{1}{(T-2)p}  \sum_{t=1}^{T-2} \| x_t-2x_{t+1} + x_{t+2}\|_1,
\EEQ
we obtain a class who entries are typically piecewise linear.
(This is discussed under the name $\ell_1$-trend filtering in 
\S\ref{s-related-work}.)
%

\paragraph{Periodic.} The component class of signals with period $P$ has loss 
function
\BEQ\label{e-periodic}
\phi(x) = \left\{ \begin{array}{ll} 0 & x_{t+P} = x_t, \quad t=1, \ldots, T-P,\\
\infty & \mbox{otherwise}. \end{array} \right.
\EEQ
We can also express this using a basis.

To this constraint we can add a loss function such as mean-square signal or 
mean-square smooth, to obtain, for example, the component class of $P$-periodic
mean-square smooth signals.  (In this case the differences are computed 
in a circular fashion.)

\paragraph{Quasi-periodic.}  A variation on the periodic signal class does not 
require strict periodicity, but allows some variation period to period, with a 
penalty for variation.
The simplest version uses the quadratic loss function
\BEQ\label{e-quasi-periodic}
\phi(x) = \sum_{t=1}^{T-P} \|x_{t+P}-x_t \|_2^2,
\EEQ
the sum of squares of the differences in signal values that are $P$ period 
apart.
Variations include adding a smoothness term, or 
replacing the sum of squares with a sum of norms, which tends 
to give intervals of time where the signal is exactly periodic.

\paragraph{Composite classes.} Components may be combined to generate more 
complex loss functions. An example that we will use later has time entries that 
are smooth~\eqref{e-ss-second} and periodic~\eqref{e-periodic} and entries that 
are mean-square close~\eqref{e-entries-close},
\BEQ\label{e-smooth-periodic-close}
\phi(x;\lambda_1,\lambda_2) = \left\{ \begin{array}{ll} 
\lambda_1 \ell_1(x) + \lambda_2 \ell_2(x)
 & x_{t+P} = x_t, \quad t=1, \ldots, T-P,\\
\infty & \mbox{otherwise}. \end{array} \right.
\EEQ
where
\BEAS
\ell_1(x) &=& \frac{1}{(T-2)p}\sum_{t=1}^{T-2}\|x_t -2x_{t+1} + x_{t+2}\|_2^2, 
\\
\ell_2(x) &=& \frac{1}{p}\sum_{t=1}^{T}\sum_{i=1}^p(x_{t,i} - \mu_t)^2,\quad 
\mu_t=\frac{1}{p}\sum_{i=1}^p x_{t,i}.
\EEAS
This composite example is convex quadratic (\S\ref{s-convex-quad}).

\paragraph{Monotone non-decreasing.}
The monotone nondecreasing loss is
\[
\phi(x) = \left\{ \begin{array}{ll}
1 & x_{t+1,i} \geq x_{t,i} \quad t=1, \ldots, T-1, \quad i=1, \ldots, p\\
0 & \mbox{otherwise}.
\end{array}\right.
\]
It is used in monotone or isotonic regression, typically to represent
something like cumulative wear, that does not decrease over time.
This loss is a constraint, but we can add an additional term such
as mean-square smoothness.

%

\paragraph{Markov.}
The Markov class is, roughly speaking, an extension of the finite set class 
\eqref{e-finite-set}
that includes costs for the different values, as well as transitions between 
them.
It is specified by some distinct values $\theta_1, \ldots, \theta_M \in 
\reals^p$,
a transition cost matrix $C \in \reals_+^{M \times M}$, and state cost vector 
$c \in \reals_+^M$.
Like the finite set component class, the loss is $\infty$ unless for each $t$, 
we have 
$x_t \in \{\theta_1, \ldots, \theta_M\}$.  
We write this as $x_t = \theta_{s_t}$, where we interpret $s_t\in \{1, \ldots, 
M\}$ as the state 
at time $t$.
When this holds, we define
\[
\phi(x) = \sum_{t=1}^T c_{s_t} + \sum_{t=2}^T C_{s_t,s_{t-1}}.
\]
The first term is the state cost, and the second is the cost of the state 
transitions.

This component class gets it name from a statistical interpretation in terms of 
a Markov
chain.  If the state $s_t$ is a Markov chain with states $\{1, \ldots, M\}$,
with transition probabilities $\pi_{ij} = \Prob(s_t = i \mid s_{t-1}=j)$.
Then with $c=0$ and $C_{i,j} =  \log \pi_{ij}$, the loss is the negative 
log-likelihood,
up to a constant.

The proximal operator of this component loss function can be efficiently 
evaluated using 
standard dynamic programming.  We create a graph with $MT$ nodes, with each 
node corresponding
to one state at one time.  
All nodes at time $t$ are connected to all nodes at time $t-1$ and $t+1$, so 
there are
$(T-1)M^2$ edges.  Let $v$ be the signal for which we wish to evaluate the 
proximal operator.
At each node we attach the cost $(\rho /2)\| v_t-c_{s} \|_2^2$, 
and on each edge from state $s$ at time $t-1$
to state $s'$ at $t$ we attach the cost $C_{s,s'}$.
Then $\phi(x) + (\rho/2) \|v-x\|_F^2$ is exactly the path cost through this 
graph. 
We can minimize this over $s_1, \ldots, s_T$ using dynamic programming to find 
the shortest 
path.  The cost is $O(TM^3)$ flops, which is linear in the signal length $T$.

%

\paragraph{Single jump.}
As a variation on the Markov component class we describe the single
jump component class.
We describe it for a scalar signal \ie, $p=1$; it is extended to vector 
signals with a loss that is separable across entries.
The loss function is
\BEQ\label{e-single-jump}
\phi(x) = \left\{ \begin{array}{ll}
1 & x = (0_\tau, a \ones_{T-\tau})\\
0 & x = 0\\
\infty & \mbox{otherwise,}
\end{array}\right.
\EEQ
for some (jump magnitude) $a \neq 0$ and some (jump time)
$\tau \in \{1,\ldots, T\}$.
Roughly speaking, feasible signals start at zero and either stay zero, or 
jump once, at a time $\tau$, to the value $a$.
The cost is zero if $x$ is zero, and one if it does jump.
This loss function is evidently nonconvex.

Its proximal operator is readily evaluated directly, by evaluating
\[
(\rho/2) \|x-v \|_2^2 + \phi(x)
\]
for all feasible $x$.  For $x=0$ we have the value $(\rho/2)\|v\|_2^2$.
For a jump at time $\tau$, the value of $a$ that minimizes the cost above
is simply the average of $x_t$ over $t = \tau, \ldots, T$.
This value and the cost is readily computed recursively, so 
the proximal operator can be evaluated in time linear in $T$.
This method extends reaadily to the masked proximal operator.

\subsection{Fitting component class losses} \label{s-fitting-losses}
In the discussion above we specify component classes directly in terms of
the loss function.  We mention here that it is also possible to fit a component
class loss from examples of signals in that class, assuming they are available.

One simple method is based on the statistical
interpretation given in \S\ref{s-stat-interp}.
Given a collection of example signals, we fit a statistical model,
for example a Gaussian distribution $\mathcal N(\mu,\Sigma)$ 
with an appropriate mean $\mu \in \reals^{Tp}$ 
and covariance $\Sigma \in \reals^{Tp \times Tp}$.
We use as loss for this component class the convex quadratic
$\phi(x) = (x_{:}-\mu)^T \Sigma^{-1} (x_{:}-\mu)$,
which is the negative log-likelihood, up to a scale factor and constant.
If we fit a statistical model for
each component of the signals we obtain an entry-separable loss; 
if we fit a common model for the entries of the signal, we obtain an 
entry-separable symmetric loss.
We can fit a time-invariant loss by creating a common statistical model
of all $M$-long slices of the signal examples, and using 
the negative log-likelihood as the slice loss.

Another elementary method for fitting a loss to example signals
uses the singular value decomposition (SVD) or generalized low-rank 
model~\cite{Udell2016}
to find a set of archetype signals $a^1, \ldots, a^r \in \reals^{T \times p}$, 
for which
each of the examples is close to a linear combination of them.  We then use the 
basis
loss function
\BEQ\label{e-basis}
\phi(x) = \left\{ \begin{array}{ll} 0 &
x = z_1a^1 + \cdots + z_r a^r ~\mbox{for some}~ z \in \reals^r\\
\infty & \mbox{otherwise}.
\end{array} \right.
\EEQ
(As a variation on this, we can find a set of (scalar) archetypes in $\reals^T$ 
for which 
each component of the examples in close to a linear combination of them, 
as in \eqref{e-comp-basis}.)
A soft version of the basis loss is the loss function
\BEQ\label{e-basis-soft}
\phi(x) = \min_{z} \|
x - z_1a^1 - \cdots - z_r a^r \|_F^2,
\EEQ
which has full domain.  (It can also be thought of as a combination of 
two classes: the basis class, and the a mean-square small residual class.)

The soft basis model can be used to fit a time-invariant loss.
We use SVD to find a set of archetypes or basis for which each $M$-long slice
of each exmaple is close to a linear combination, and then use 
the soft basis loss \eqref{e-basis-soft} as the slice loss.

\section{Examples} \label{s-examples}

\subsection{Mauna Loa CO$_2$ measurements} \label{s-CO2}
An example often used to demonstrate seasonal-trend decomposition is
atmospheric carbon dioxide (CO$_2$), which has both a strong seasonal 
component and a underlying trend. 
These data were utilized in the original STL paper~\cite{Cleveland1990} as 
well as the documentation for various implementations of 
STL~\cite{stl-python}. 
In this section we compare the 
Python implementation of STL in the \texttt{statsmodels} package to an 
SD formulation of the problem 
of decomposing measurements of atmospheric CO$_2$ into seasonal, trend, 
and residual components. 

\begin{figure}
\centering
\centering
\resizebox{0.95\columnwidth}{!}{
\import{figs/}{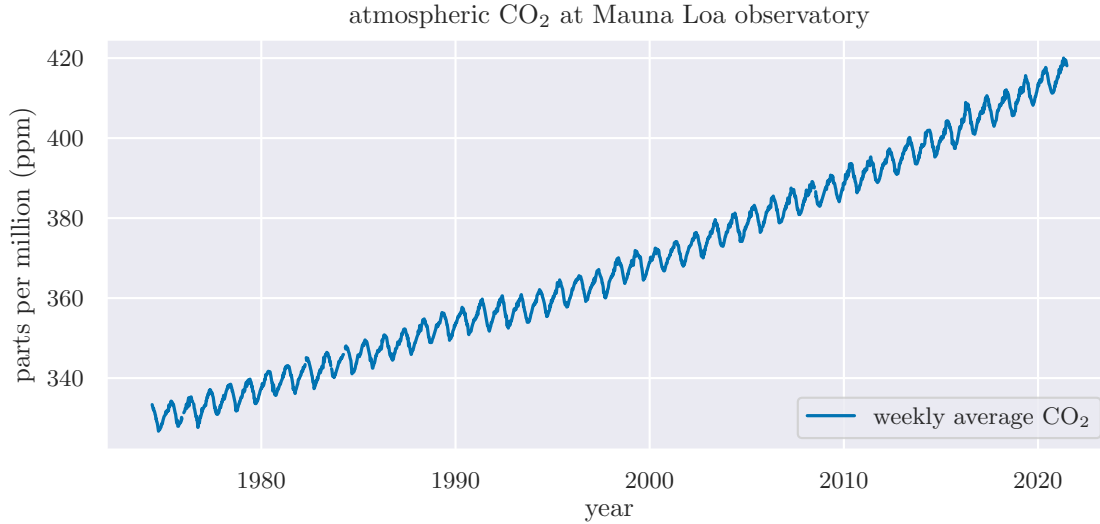}
}
\caption{Atmospheric CO$_2$ data obtained from NOAA, which shows clear seasonal 
and trend
components.}
\label{f-co2-data}
\end{figure}

\paragraph{Data set.} The weekly average CO$_2$ measured at Mauna Loa, HI from 
May 1974 through 
June 2021, available online from the National Oceanic and Atmospheric 
Administration Global 
Monitoring Laboratory~\cite{co2-data}, is shown in figure~\ref{f-co2-data}. 
The data set is a scalar signal of length 2459 with 18 missing entries.
In our notation, $y_1,\ldots,y_T\in\reals\cup \{?\}$, with $T=2459$, and 
$\left\lvert \mathcal U\right\rvert = 18$.

\begin{figure}
\centering
\resizebox{0.95\columnwidth}{!}{
\import{figs/}{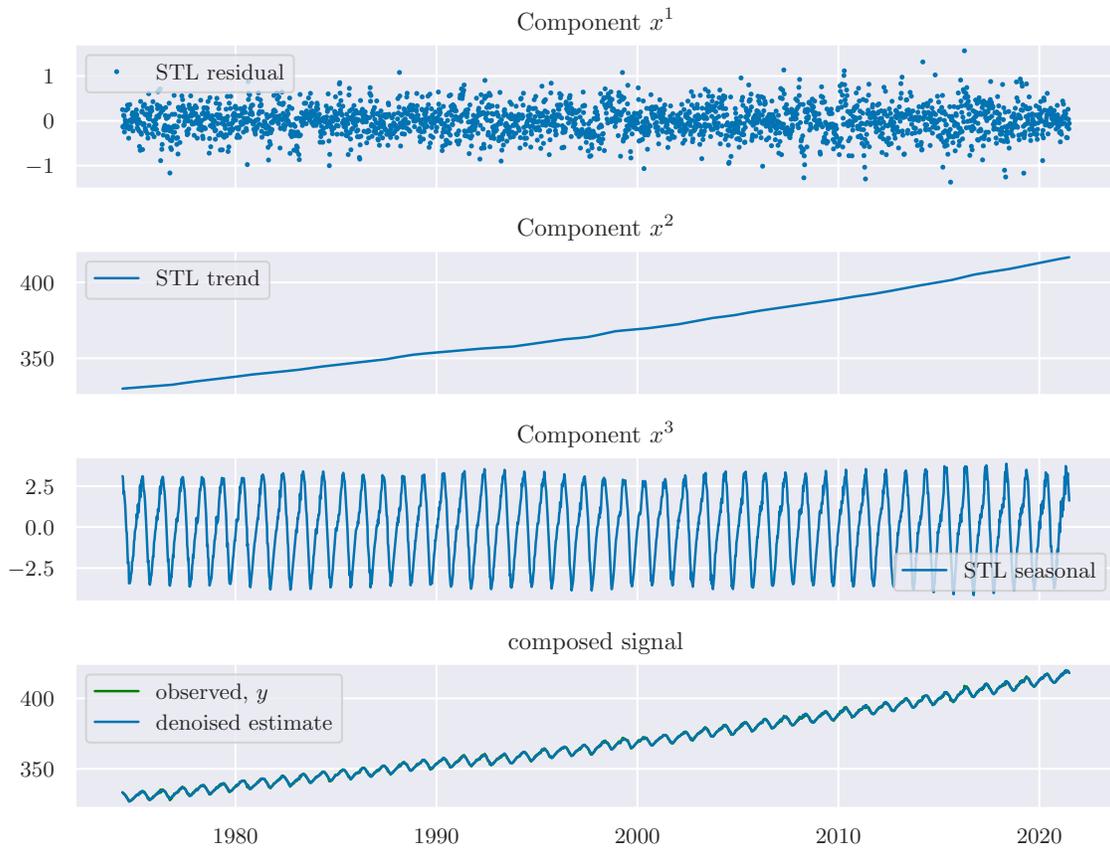}
}
\caption{Decomposition of the CO$_2$ data into residual, trend, and 
seasonal components, using STL.}
\label{f-co2-stl}
\end{figure}

\paragraph{Decomposition using STL.}  We use the implementation in 
\texttt{statsmodels} (v0.12.2) with default 
settings and \texttt{period=52}. We note that while the original STL paper 
describes
how to handle missing data, this particular software implementation cannot 
handle missing values,
so we used simple linear interpolation to fill the missing values 
before running the algorithm. The resulting decomposition is shown in 
figure~\ref{f-co2-stl}, using the 
conventional names for the components. 
Interestingly, the ``seasonal'' component in this estimation is 
not periodic; it almost repeats each year but with some variation.

\begin{figure}
\centering
\resizebox{0.95\columnwidth}{!}{
\import{figs/}{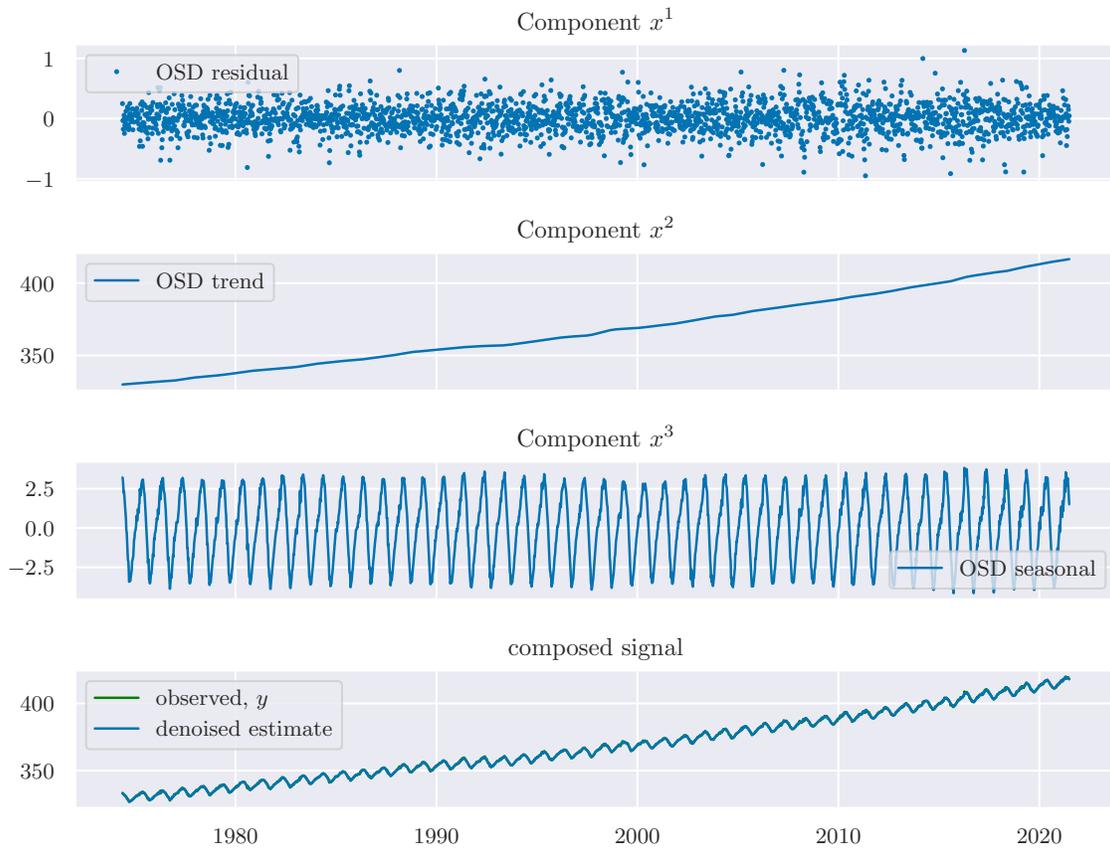}
}
\caption{Decomposition of the CO$_2$ data into 
residual, trend, and seasonal components, using SD.
It is nearly identical to the decomposition
found by STL, shown in figure~\ref{f-co2-stl}.}
\label{f-co2-osd}
\end{figure}

\paragraph{Decomposition using SD.} 
We form an SD problem with $p=1$, $T=2459$, and $K=3$, with
component classes
mean-square small~\eqref{e-ms-small}, second-order-difference 
small~\eqref{e-ss-second}, and a 
quasi-periodic signal with period $52$~\eqref{e-quasi-periodic}.  All the 
component classes are 
convex, so this SD problem is convex. This problem has two parameters 
$\lambda_2$ and $\lambda_3$, 
associated with the weights on the second and third loss functions 
respectively. We found that 
$\lambda_2=10^4$ and $\lambda_3=1$ give good results, although better parameter 
values could be 
found using a validation procedure. The resulting decomposition is shown in 
figure~\ref{f-co2-osd}. 

\paragraph{Comparison.} 
The decompositions found using STL and SD, shown in figures~\ref{f-co2-stl} 
and \ref{f-co2-osd}, and nearly identical.
The RMS deviation between trend estimates is $7.52\times10^{-2}$, about 
$0.02\%$ of the 
average measured value. The RMS deviation between seasonal estimates is 
$8.79\times 10^{-2}$. 
While STL is based on a heuristic algorithm, SD is based on solving a convex 
optimization problem 
(for our particular choice of loss functions).

\clearpage
\subsection{RFK bridge traffic}\label{s-traffic}
This example illustrates how the concept of seasonal-trend decomposition can 
be extended in the SD framework to handle more complex analyses with
additional components.
Traffic volume is measured with sensors embedded in the 
roadways that count the number of cars that pass in each hour;
from these data, 
summary statistics such as ``Annual Average Daily Traffic'' and 
``Peak Hour Volume'' are derived~\cite{traffic-census}. 

\begin{figure}
\centering
\resizebox{\columnwidth}{!}{
\import{figs/}{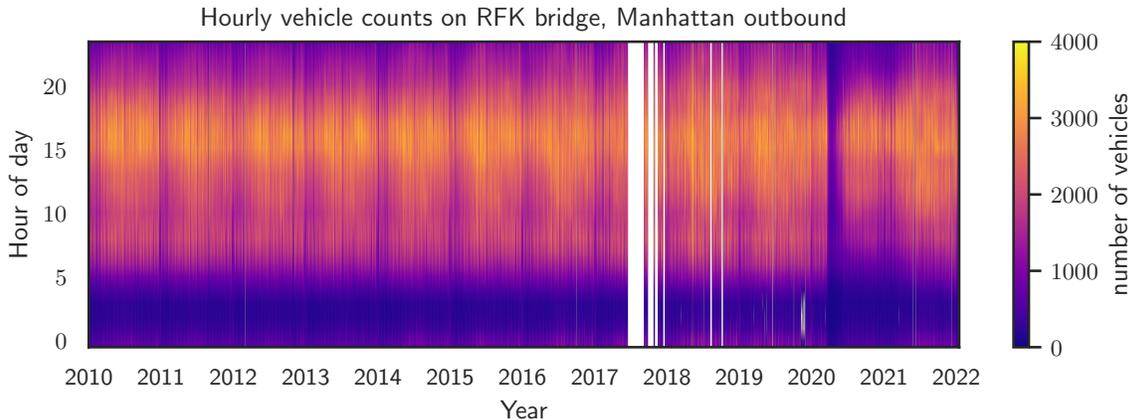}
}
\caption{Hourly vehicle counts for the outbound Manhattan toll plaza of the 
Robert F. Kennedy 
bridge, with hour of day on the y-axis and days on the x-axis. 
White pixels represent missing values. 
}
\label{f-traffic-data}
\end{figure}

\begin{figure}
\centering
\resizebox{0.9\columnwidth}{!}{
\import{figs/}{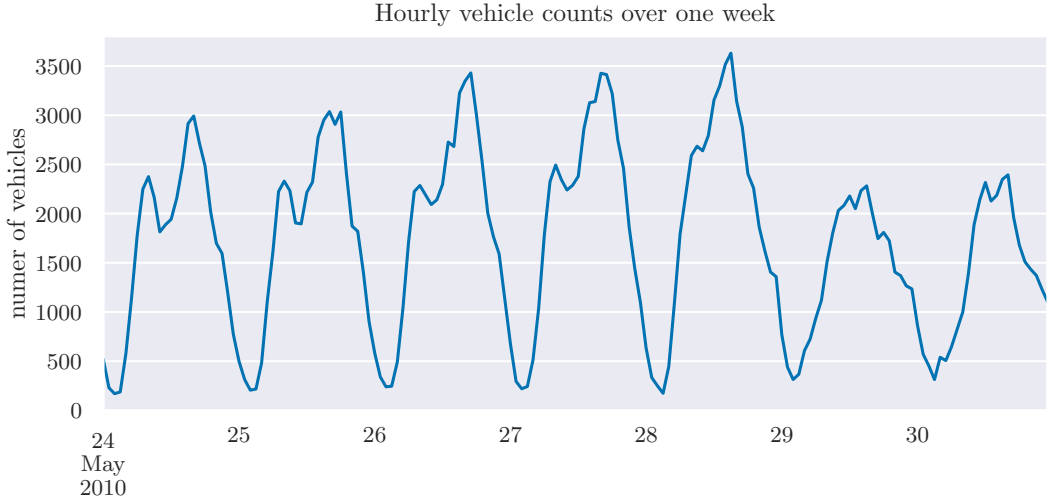}
}
\caption{One week of hourly vehicle counts in May 2010.}
\label{f-traffic-week}
\end{figure}

\paragraph{Data set.}
The hourly outbound vehicle count for the Manhattan toll plaza on the Robert F. 
Kennedy Bridge in 
New York City from January 1, 2010 through August 28, 2021 is shown in 
figure~\ref{f-traffic-data}
as a heat map,
with the hour of day shown vertically and the day shown horizontally, and 
missing entries shown 
in white.
Daily and seasonal variations can be seen, along with the effects of COVID-19.
A single week of data is shown in figure~\ref{f-traffic-week}, where daily 
variation,
and the weekend effect, are evident.
The data set is made 
available online by the New York Metropolitan Transportation Authority 
(MTA)~\cite{traffic-data}. 

The data $y$ is scalar (\ie, $p=1$), with $T=102192$ (24 hours per day, 4258 
days). 
We take the natural logarithm of the data, using the 
convention $\log 0 =?$. With these unknown entries, plus those that are unknown 
in the original
data set, we have $\left\lvert \mathcal U\right\rvert = 3767$.
Thus the decomposition is multiplicative; the components are multiplied to 
obtain the decomposition.


\paragraph{SD problem formulation.}

We form an SD problem with $K=5$ components. The residual component is 
mean-square 
small~\eqref{e-ms-small}, as in previous examples. The second component is the 
weekly
baseline, which is the smooth-periodic cost given in 
(\ref{e-P-periodic-smooth}) with $P=168$ and a 
weight parameter $\lambda_2$. The third component is the yearly seasonal 
correction, which is 
also smooth-periodic (\ref{e-P-periodic-smooth}) with $P=8760$ and weight 
parameter 
$\lambda_3$, and the additional constraint that the sum over each period 
must be equal to zero. The 
fourth component is the long-term trend, modeled as piecewise 
linear with the $\ell_1$ second difference loss~(\ref{e-l1-trend}),
with weight parameter $\lambda_4$ and the additional constraint that the 
first value of $x^4$ must be equal to zero. The fifth and final component is a 
sparse daily outlier, defined as 
\BEQ\label{e-daily-sparse}
\phi_5(x) = \left\{ \begin{array}{ll}
\lambda_5 \| x\|_1 & x \in \mathcal D\\
\infty & \mbox{otherwise},
\end{array}\right.
\EEQ
where $\mathcal D$ is the set of signals that are constant over each day.
All the component class losses are 
convex, so this SD problem is convex with parameters $\lambda_2$, $\lambda_3$, 
$\lambda_4$, 
and $\lambda_5$.

\paragraph{Results.} We solve the SD problem using parameter values
\[
\lambda_2 = 10^{-1}, \quad
\lambda_3 = 5\times 10^5, \quad
\lambda_4 = 2\times 10^5, \quad
\lambda_5 = 1,
\]
selected by hand to provide good results. The decomposition yields components 
that have vastly 
different timescales. 

\begin{figure}
\centering
\resizebox{0.6\columnwidth}{!}{
\import{figs/}{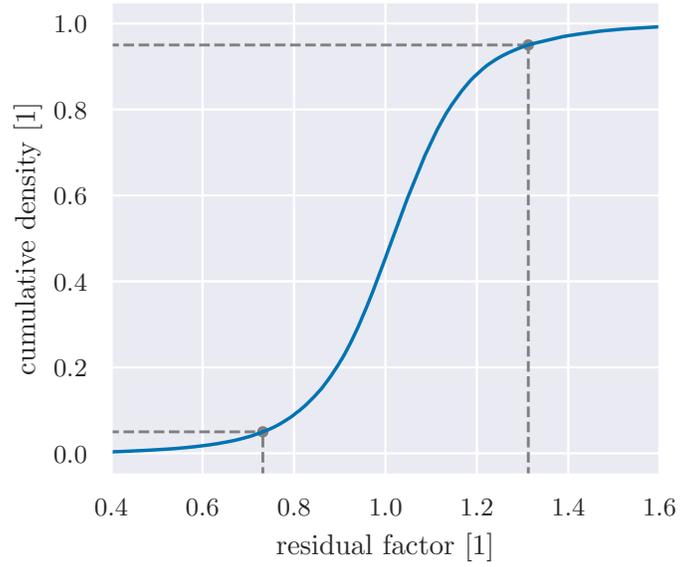}
}
\caption{Cumulative distribution function of the multiplicative residual
$\tilde{x}^1_{t,i}$ for $(t,i)\in\mathcal K$. The gray dashed lines indicate 
the 
5th and 95th percentiles. 90\% of the residuals 
are between 0.74 and 1.30.}
\label{f-traffic-residual}
\end{figure}

By exponentiating the component estimates,
$\tilde{x}^k=\exp(x^k)$, we recover a multiplicative model of the traffic count 
data. 
The residual component $\tilde{x}^1$ is centered around 1, with 90\% of the 
residuals 
in the interval $[0.74,1.30]$, shown in figure~\ref{f-traffic-residual}. 
This means that in any given hour, the decomposition predicts traffic typically
within around $\pm 30\%$.

\begin{figure}
\centering
\resizebox{\columnwidth}{!}{
\import{figs/}{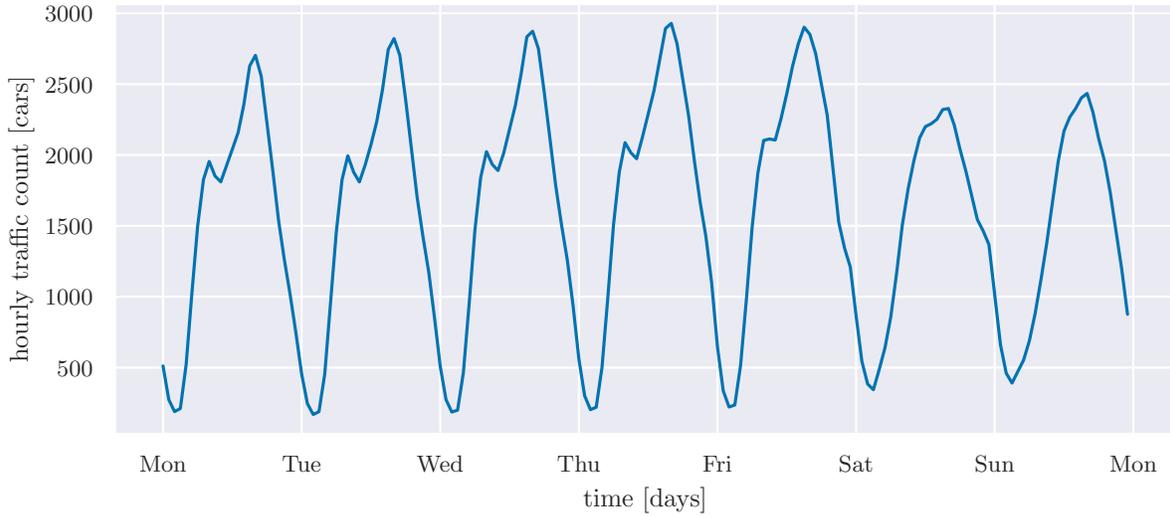}
}
\caption{Weekly baseline signal $\tilde{x}^2$, shown for a single week (168 
values).}
\label{f-traffic-avg-week}
\end{figure}

Figure~\ref{f-traffic-avg-week} shows one week of the (periodic) weekly 
baseline.
We see many of the phenomena present in figure~\ref{f-traffic-week}, such as 
reduced traffic over the weekend, daily variation, and a small increase from 
Monday to Friday, and a commute rush hour on weekdays.

\begin{figure}
\centering
\resizebox{0.95\columnwidth}{!}{
\import{figs/}{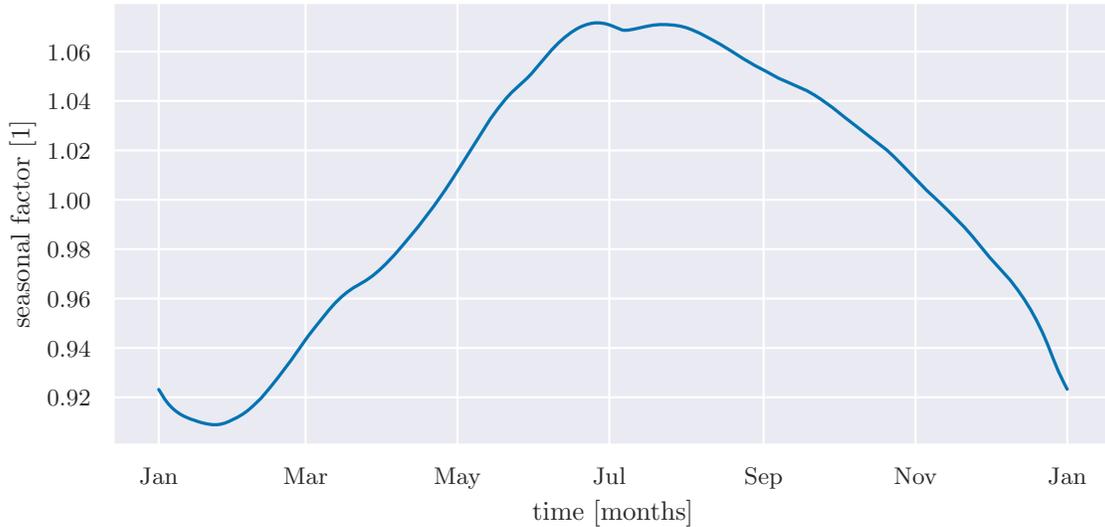}
}
\caption{Seasonal adjustment $\tilde{x}^3$, shown for a single year (8760 
values).}
\label{f-traffic-seasonal}
\end{figure}

Figure~\ref{f-traffic-seasonal} shows
component $\tilde{x}^3$, the seasonal correction factor,
which varies from around $-9\%$ to $+7\%$, with the peak in summer and
the low point in late January and early February.

\begin{figure}
\centering
\resizebox{0.95\columnwidth}{!}{
\import{figs/}{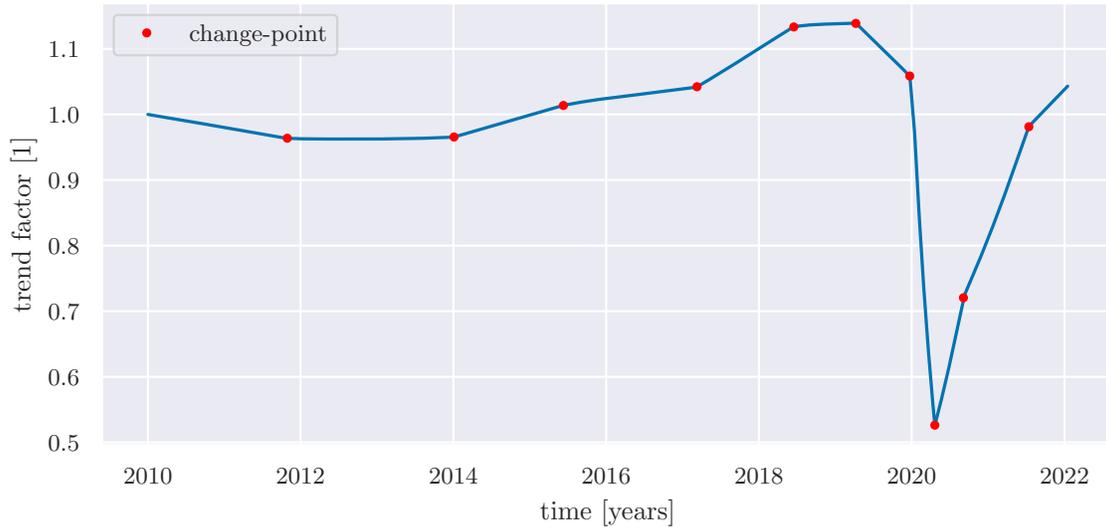}
}
\caption{Long-term trend multiplicative factor $\tilde{x}^4$. This trend is 
piecewise exponential, with breakpoints shown as red dots.}
\label{f-traffic-trend}
\end{figure}

Figure~\ref{f-traffic-trend} shows the long term
$\tilde{x}^4$.  The component $x^4$ is piecewise-linear 
with a small number of breakpoints, so $\tilde x^4$ is piecewise exponential,
with a small number of breakpoints, shown as red dots in the plot.
We can see a slight increase in traffic over the first 10 years followed by 
the a precipitous drop in 
traffic due to COVID-19 in early 2020, 
coinciding with the mandatory lockdown implemented by the 
New York state government on March 22, 2020~\cite{ny-lockdown}. 

\begin{figure}
\centering
\resizebox{0.95\columnwidth}{!}{
\import{figs/}{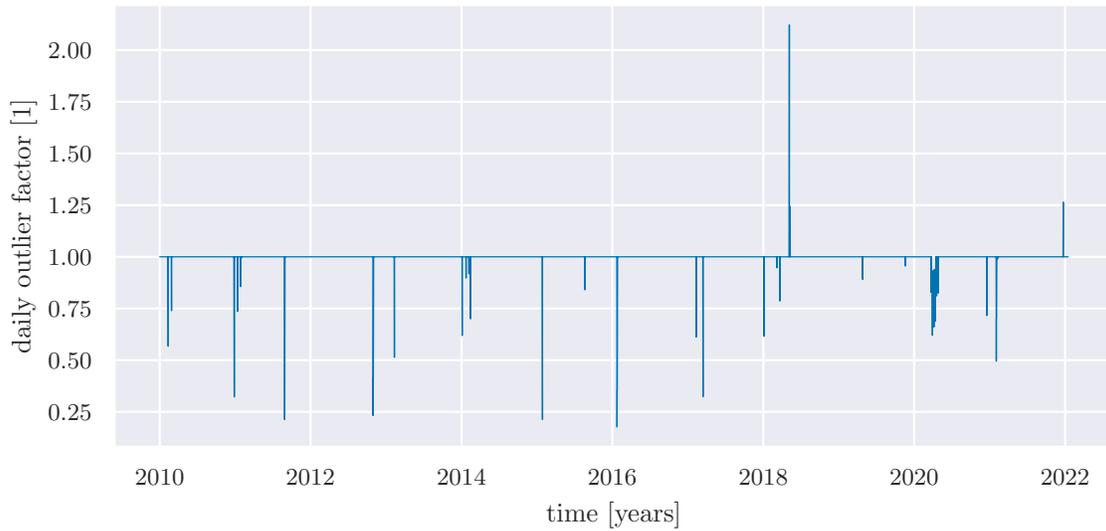}
}
\caption{Daily outlier component $\tilde x^5$.}
\label{f-traffic-outliers}
\end{figure}

\begin{figure}
\centering
\resizebox{\columnwidth}{!}{
\import{figs/}{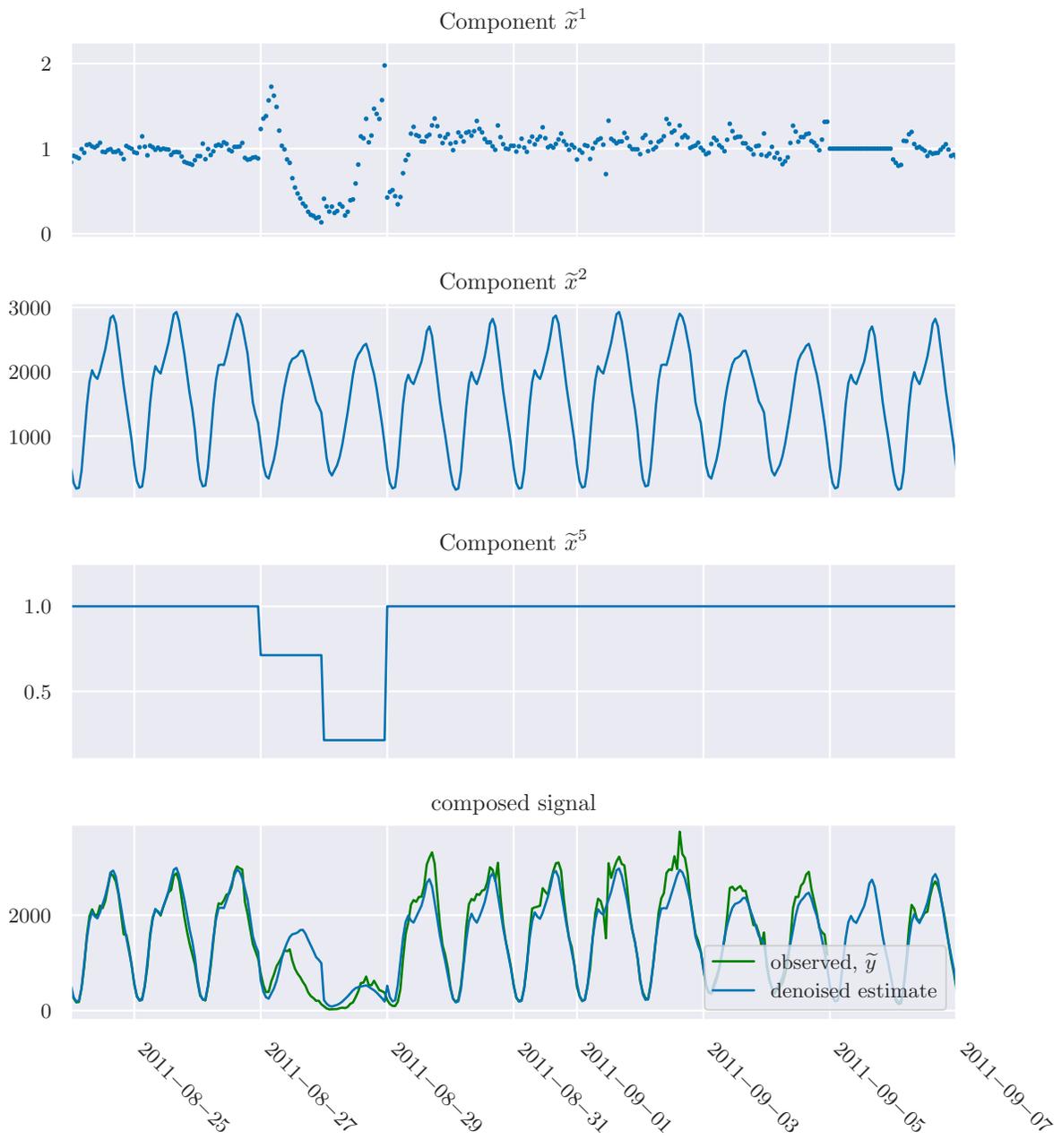}
}
\caption{Decomposition components for two weeks in August 2011 (336 values).
Hurricane Irene hit New York city on August 27 and 28, 
greating reducing traffic on those days, clearly seen as outliers in $\tilde 
x^5$.}
\label{f-traffic-decompose-short}
\end{figure}

The final component $x^5$ is sparse, which means that $\tilde x^5$,
shown in figure~\ref{f-traffic-outliers}, mostly takes on the value one.
This component identifies 42 days (out of 4258) as outliers, with 
multiplicative 
corrections ranging from around $0.2$ (\ie, one fifth the normal traffic 
on that day) to around twice the normal traffic on that (one) day.
All but two of the outliers represent a decrease in traffic on that day.
Many of the detected outlier days are 
weather related, with some notable examples being various blizzards including 
February 10, 
2010~\cite{blizzard2010}, December 27, 2010~\cite{blizzard2010b}, January 27, 
2015~\cite{blizzard2015}, and February 1, 2021~\cite{blizzard2021}.
About 9 outlier days are associated with reduced traffic 
during the COVID-19 lockdown event in early 2020. 
Figure~\ref{f-traffic-decompose-short} highlights the detection of 
Hurricane Irene in August of 2011~\cite{Avila2011}, with $\tilde x^5<1$ during
the hurricane. 

The two positive outlier days occur on May 6 and 10, 2018. 
The authors could find no explanation for the very 
high measured traffic on those days in the archives of the New York Times and 
the New York Post.
It is possible that sensors were simply malfunctioning on those two days.

\clearpage
\subsection{Outage detection in a photovoltaic combiner box}\label{s-pv-fleet}

\paragraph{Data set.}
We consider a set of 7 measurements of real power from inside a photovoltaic 
(PV) combiner 
box~\cite{Franklin2018}, corresponding to 7 strings of series-connected PV 
modules 
that are joined in parallel. These data are from PV strings forming the canopy 
at the 
NIST campus in Maryland~\cite{NIST}. 
Detailed documentation of the PV systems at this site, including system 
designs, 
meteorological station information, and site layout, are also 
available~\cite{NISTdoc}. 
The canopy has multiple roof orientations, so the constituent 
strings have similar but different power curves, depending on the 
specific geometry of each string.

\begin{figure}
\centering
\resizebox{\columnwidth}{!}{
\import{figs/}{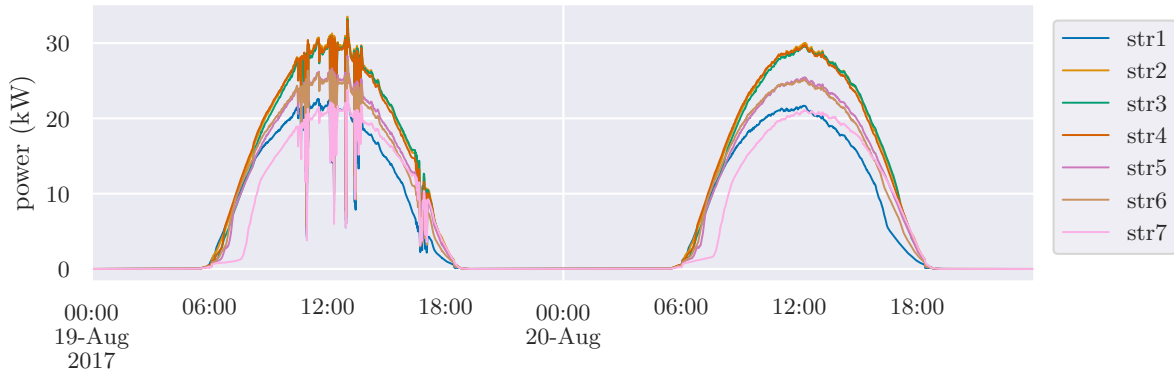}
}
\caption{Raw PV combiner box data, shown for two days in August 2017.}
\label{f-pv-example}
\end{figure}

The raw data consist of the power output of each of the 7 PV strings, 
measured each minute over a month (August 2016), organized into a matrix
with each column corresponding to a single string and each row a particular 
minute. 
This raw data contains some missing data.
The power output of each string depends on available sunlight, weather 
conditions, 
soiling accumulation, string geometry, and 
local shade patterns. Two days of string power output are shown in 
figure~\ref{f-pv-example}.

\paragraph{Data pre-processing.}
We first eliminate all data points corresponding to night time and early 
morning 
and evening, when string powers are zero or very small. 
We removed data between 5:40pm and 6:49am.  (These times were found as the times
when the whole system was producing less than 10\% of system capacity.)
Thus each day consists of 652 one minute measurements.
Next we scale each of the 7 string powers (columns) so that the 95th percentile 
is one.
This gives each column an approximate range of about 0.1 to 1.3. 

Finally we take the log of each power output value,
resulting in columns with a range of about -2.3 to 0.25.
Carrying out signal decomposition on this log signal gives us a 
multiplicative decomposition, which makes sense for this application.
(For example, a cloud passing between the sun and the string gives a 
percentage reduction in power.)
The final data is a signal~$y$ with 
$T=20212$, $p=7$, and $|\mathcal{U}| = 6402$. 

\begin{figure}
\centering
\resizebox{\columnwidth}{!}{
\import{figs/}{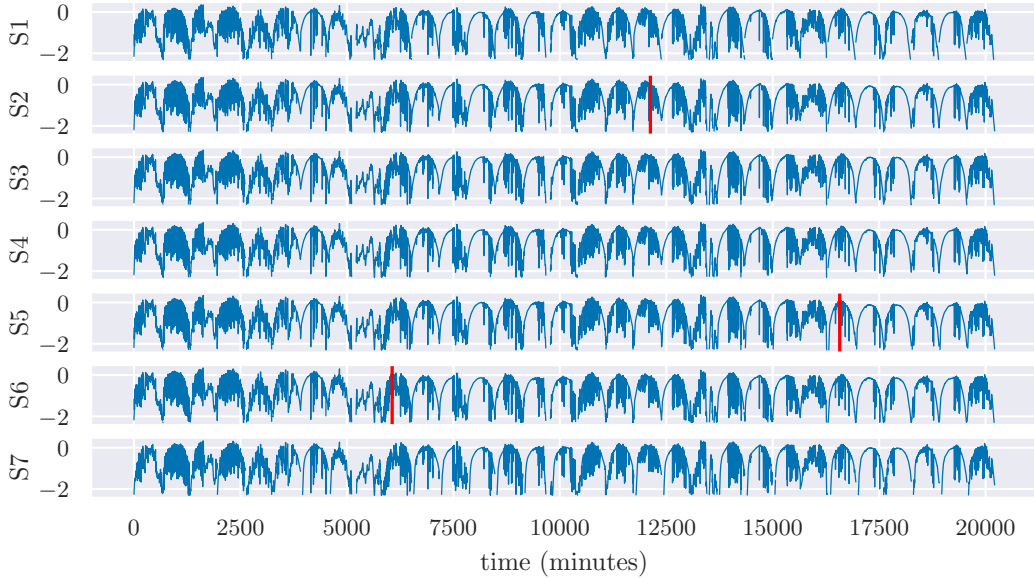}
}
\caption{PV combiner box data after pre-processing, with simulated outages.
The onset times of the simulated outages are shown as vertical red lines.}
\label{f-pv-fleet}
\end{figure}

\paragraph{Outage simulation.}
We modify this real data to include some simulated faults or outages,
where some part of each PV string no longer generates power.
This is modeled as a (multiplicative) reduction 
in power output, from the time of failure to the end of the data.
We simulated these fault for strings $2$, $5$, and $6$,
with onset times 
\[
T_2 = 12132, \quad
T_5 = 16573, \quad
T_6 = 6063,
\]
and power reduction factors
\[
f_2 = -7\%, \quad
f_5 = -10\%, \quad
f_6 = -12.5\%,
\]
chosen randomly.  (These are realistic values.)
The modified data is shown in figure~\ref{f-pv-fleet}, with 
vertical red lines indicating the onset of the outages in spower trings $2$, 
$5$, and $6$.
These power reductions can be seen in the plot, but would likely be hard to 
spot by eye.

\paragraph{SD problem formulation.} 
We form an SD problem with $K=5$ components. Our signal 
decomposition models string output as the product of a mean-square small 
residual \eqref{e-ms-small}, a
clear sky signal, a common daily correction term, a common cloud/weather term, 
and a failure
term.
The clear sky component is modeled as the composite class that is smooth and 
periodic in time and close in entries~\eqref{e-smooth-periodic-close} (with a 
small modification to remove the smoothness penalty across day boundaries).
This component has two parameters,
one for the smoothness term and one for the variance across entries, 
$\lambda_{2a}$ and 
$\lambda_{2b}$, respectively. 
The third component is a daily scale 
adjustment that is constant across columns, and constant over each day,
meant to capture day-to-day macro-scale changes in atmospheric 
conditions that effect all strings, such as precipitable water and aerosol 
optical 
depth~\cite{Ineichen2008}. 
The fourth component is also constant across the columns and has a quantile 
loss 
function~\eqref{e-quantile}. This 
models a common cloud-loss term between the strings, assumed to be equal 
because the 
strings are so close to each other and are experiencing the same local weather. 
The fourth component 
has two parameters, the quantile term, $\tau$, which we set to be $0.65$, and a 
weight, $\lambda_4$. 
The third and fourth 
components make use of the common term 
formulation~\eqref{e-common-term}. The fifth component is the failure detector. 
This component 
uses the single jump class~\eqref{e-single-jump}, constrained to only have 
negative jumps, with each 
column treated independently. The fifth component also has a weight parameter, 
$\lambda_5$.
Since the failures are simulated, we know exactly when the onsets are, and 
what the values
are, which we can compare to the estimated failure component.

\begin{figure}
\centering
\resizebox{\columnwidth}{!}{
\import{figs/}{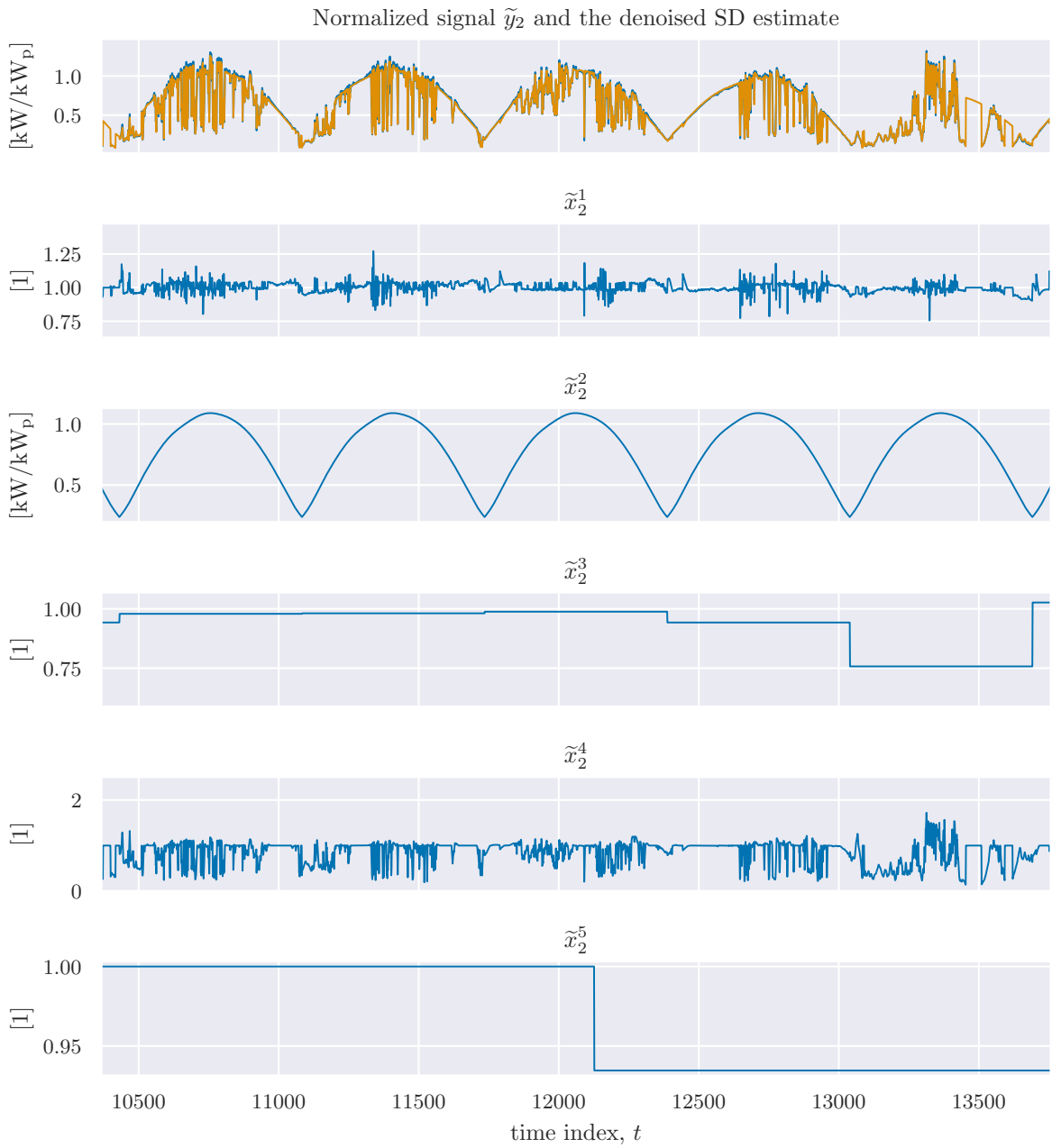}
}
\caption{Components $\tilde{x^k}$ for string 2 over 5 days.}
\label{f-pv-sys1-decomp}
\end{figure}

\paragraph{Results.} We solve the SD problem with hand-selected weights,
\[
\lambda_{2a} = 5\times10^{4}/(Tp), \quad
\lambda_{2b} = 5\times10^{-5}/(Tp), \quad
\lambda_4 = 2/(Tp), \quad
\lambda_5 = 10/(Tp),
\]
giving us estimates of $x^1, \ldots, x^5$. 
Our estimates of the components are $\tilde{x^k} = \exp x^k$.  We interpret 
$\tilde x^1,\tilde x^2,
\tilde x^4, \tilde x^5$ as multiplicative components, and we interpret $\tilde 
x^2$ as
the baseline clear sky values, normalized.
It takes approximately 15 seconds to run the 
SD-ADMM algorithm to convergence on a 2016 MacBook Pro, with no parallelization 
of the 
proximal operator evaluations. A segment of the decomposition is shown in 
figure~\ref{f-pv-sys1-decomp}, highlighting 5 days of data for string 2, 
including 
the time of an estimated failure.

\begin{figure}
\centering
\resizebox{0.65\columnwidth}{!}{
\import{figs/}{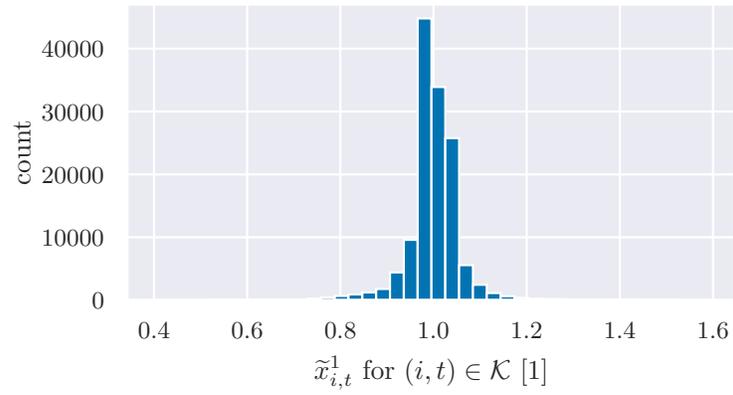}
}
\caption{Histogram of the residual term $\tilde{x}^1$ for all entries of the 
known set 
$\mathcal{K}$.}
\label{f-pv-residual}
\end{figure}

The residual term $\tilde x^1$ is shown as a histogram in 
figure~\ref{f-pv-residual}. The residual is 
centered at 1 and has a standard deviation of $0.082$. 95\% of the entries in 
the known set have 
residuals in the range of $[0.85,1.15]$, \ie, $\pm15\%$. 

\begin{figure}
\centering
\resizebox{0.9\columnwidth}{!}{
\import{figs/}{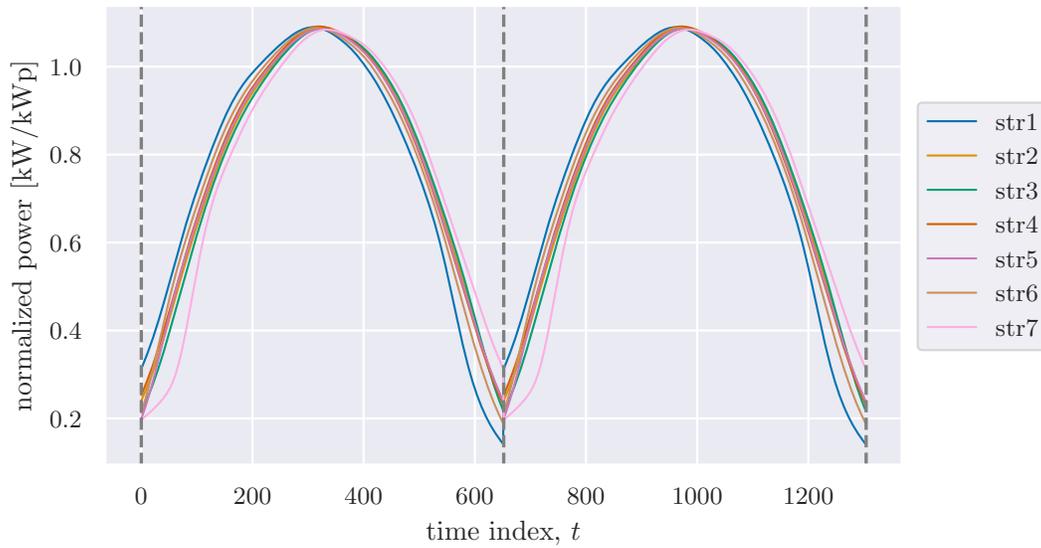}
}
\caption{The clear sky component $\tilde x^2$, with two days shown.}
\label{f-pv-clear}
\end{figure}

The clear sky component $\tilde x^2$ is shown in figure~\ref{f-pv-clear}. We 
plot two days of this 
periodic component to illustrate the discontinuities in values between adjacent 
days.
We see that the clear sky estimates for the strings are smooth in time, and 
vary a bit 
between strings. 

\begin{figure}
\centering
\resizebox{\columnwidth}{!}{
\import{figs/}{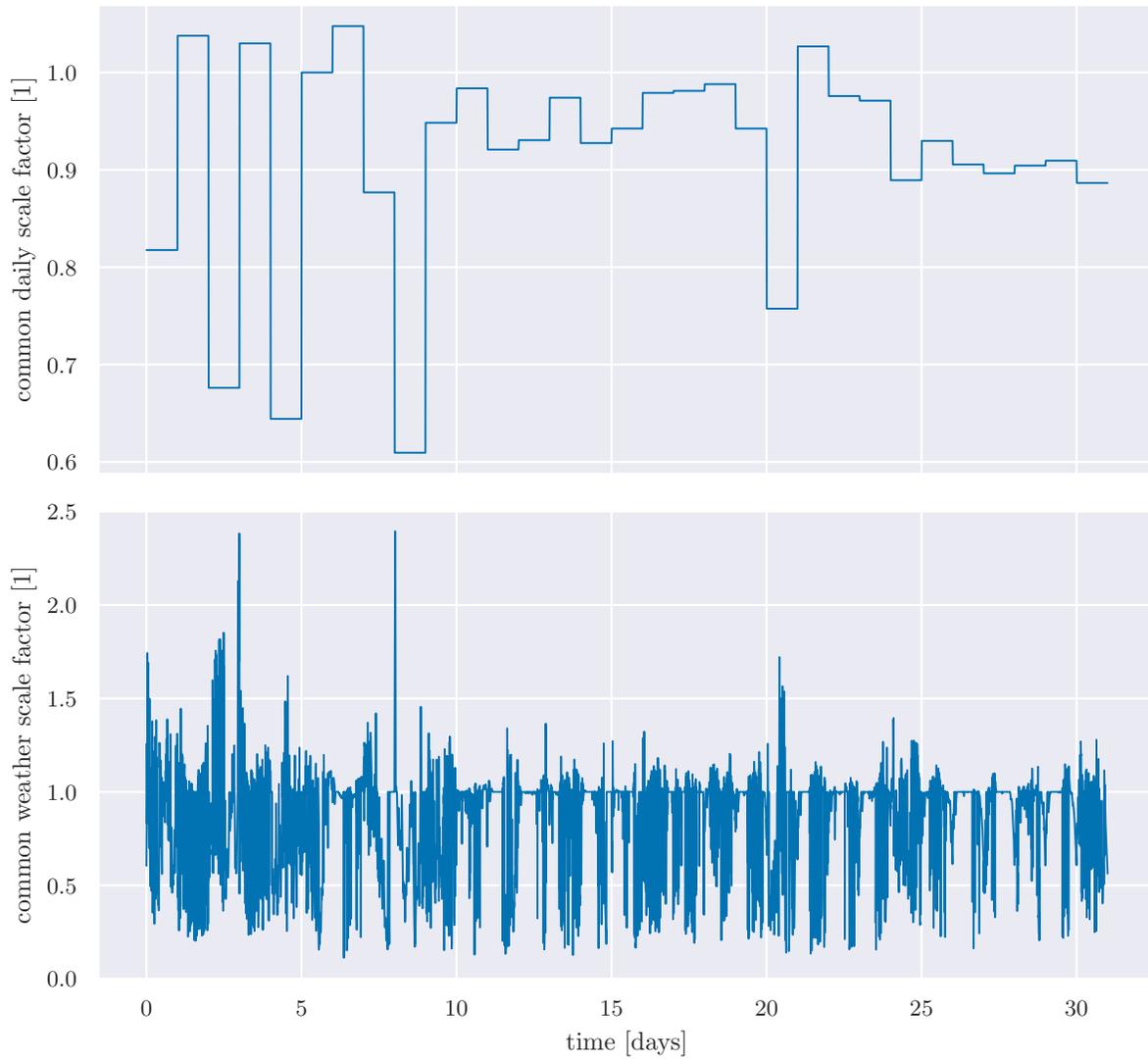}
}
\caption{(a) Top, the common daily scale factor $\tilde x^3$. (b) Bottom, the 
common weather 
component $\tilde x^4$. Only one column of each component is plotted as all 
columns are equal to 
each other.}
\label{f-pv-scale-weather}
\end{figure}

The common daily scale factor $\tilde x^3$, shown in 
figure~\ref{f-pv-scale-weather}(a), is constant 
across 
days and across columns. This can be thought of how much the clear sky signals 
need to be scaled 
to recreate any given day, and the all strings must agree on the factor. Days 
with significant cloud 
cover tend to have much smaller scale factors, while 
clearer days tend to vary by about 10--15\%.

The common weather term $\tilde x^4$, shown in 
figure~\ref{f-pv-scale-weather}(b), is also constant 
across 
columns, and it captures the effects of local weather, particularly attenuation 
by clouds. This term is 
typically a loss, 
that is $\tilde x^4 < 1$. We chose the value of the quantile parameter 
$\tau=0.65$ through hand-tuning 
and selecting a value that gave good agreement between the clear sky component 
and the measured 
data on periods without significant cloud impacts. While having a weather 
correction term that is larger 
than about 1.5 does not 
make much physical sense (see, for example,~\cite{Inman2016}), we observe that 
this factor is applied 
to the combination of components 2 and 3, the clear sky component and the daily 
scale factor. In fact, 
we see the larger values in $\tilde x^4$ exactly on the days that are highly 
cloudy and use very small 
daily scale factors.

\begin{figure}
\centering
\resizebox{\columnwidth}{!}{
\import{figs/}{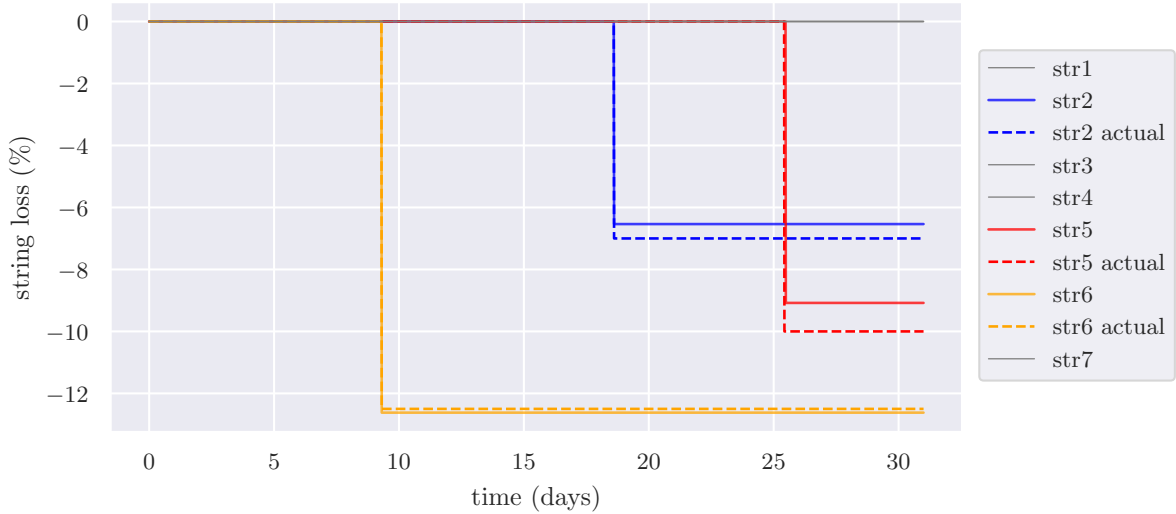}
}
\caption{Failure component, shown as the percentage $100\times (1-\tilde 
x^5)\%$. The  
dashed lines show the actual simulated failures.}
\label{f-pv-failure-detect}
\end{figure}

The failure component $\tilde x^5$ correctly identifies the failures correctly
as appearing in only strings 2, 5, and 6, as
depicted in figure~\ref{f-pv-failure-detect}, which shows the predicted and 
real failure onset times and amounts.
The estimated failure time is 5 minutes late for string 2, about 2 hours late 
for string 5, 
and exactly correct for string 6;
for all three strings, the loss was detected within the same day that the 
failure occurred.
We can also see that the estimated failure amounts are quite good.
\begin{table}
\centering
\caption{Outage detection results}
\begin{tabular}{c|c|c|c}
string & metric& actual & predicted \\
\hline
2 &  amount (\%) & -7 & -6.24 \\
2 &  time (days) & 18.60 & 18.61 \\
\hline
5 &  amount (\%) & -10 & -9.10 \\
5 &  time (days) & 25.42 & 25.60 \\
\hline
6 &  amount (\%) & -12.5 & -12.44 \\
6 &  time (days) & 9.30 & 9.30 \\
\end{tabular}
\label{t-pv-failure-detect}
\end{table}

\section*{Acknowledgments}
This material is based on work supported by the U.S. Department of Energy's 
Office of Energy Efficiency and Renewable Energy (EERE) under the Solar Energy 
Technologies Office Award Number 38529. 
This research was partially supported by ACCESS (AI Chip Center for Emerging 
Smart Systems), sponsored by InnoHK funding, Hong Kong SAR.
The authors thank Joel Tropp for useful suggestions on an early draft of this
paper.

\clearpage

\bibliography{OSD_bib}

\clearpage
\appendix

\section{SD-ADMM algorithm derivation}\label{s-sd-admm-deriv}

To derive an ADMM algorithm for SD,
we introduce new variables $z^1, \ldots, z^K \in \reals^q$ and 
reformulate the SD problem \eqref{e-sd-no-x1} as
\[
\begin{array}{ll}
\mbox{minimize}   &  \phi_1(x^1) + \cdots +
\phi_K(x^K) \\
\mbox{subject to} & \mask x^k - z^k = 0,\quad k=1,\ldots,K\\
& \mask y = z^1 + \cdots + z^K.
\end{array}
\]
We let $g$ denote the indicator function of the last constraint,
\[
g(z^1,\ldots,z^K)=\left\{ \begin{array}{ll}
0 & \mask y = z^1 + \cdots + z^K\\
\infty & \mbox{otherwise},
\end{array}\right.
\]
so the SD problem can be expressed as
\BEQ\label{e-sd-z}
\begin{array}{ll}
\mbox{minimize}   &  \phi_1(x^1) + \cdots + \phi_K(x^K) + g(z^1,\ldots,z^K)\\
\mbox{subject to} & \mask x^k - z^k = 0,\quad k=1,\ldots,K.
\end{array}
\EEQ
We write this in compact form as
\BEQ
\begin{array}{ll}
\mbox{minimize}   &  \phi(x) + g(z)\\
\mbox{subject to} & \mask x^k - z^k = 0,\quad k=1,\ldots,K,
\end{array}
\EEQ
where $x=(x^1,\ldots,x^K)$, $z=(z^1,\ldots, z^K)$, and
$\phi(x) = \phi_1(x^1) + \cdots + \phi_K(x^K)$.
We are now ready to derive the ADMM algorithm.

We form the augmented Lagrangian, with parameter $\rho>0$,
\BEAS
L_\rho(x,z,\lambda) &=& \phi(x)+g(z)+\sum_{k=1}^K\left(
{\lambda^k}^T(\mask x^k - z^k) 
+ (\rho / 2)\|\mask x^k - z^k\|_2^2\right)\\
&=& \phi(x)+g(z)+(\rho/2) \sum_{k=1}^K\left(\|r^k + u^k\|_2^2 - 
\|u^k\|_2^2\right),
\EEAS
where $r^k=\mask x^k - z^k$ are the residuals, 
$\lambda^k$ are the dual variables, and
$u^k=(1 / \rho)\lambda^k$ are the so-called scaled dual variables \cite[\S 
3.1.1]{Boyd2011}.

Iteration $j$ of ADMM consists of three steps:
\BEAS
x^{j+1} &=& \argmin_{x} L_\rho (x^j,z^j,u^j)\\
z^{j+1} &=& \argmin_{z} L_\rho (x^{j+1},z^j,u^j)\\
(u^k)^{j+1} &=& (u^k)^j + 
\mask (x^k)^{j+1} -(z^k)^{j+1}, \quad k=1, \ldots, K,
\EEAS
which we refer to as the $x$-update, $z$-update, and $u$-update, respectively.

We now work out and simplify these steps.
Since $L_\rho$ is separable in $x^k$, we can minimize over $x^k$ separately 
in the $x$-update to obtain
\BEQ\label{e-x-update1}
(x^k)^{j+1} = \argmin_{x^k}\left(\phi_k(x^k) + (\rho/2)\|\mask x^k - (z^k)^j 
+ (u^k)^j\right\|_2^2) , \quad k=1, \ldots, K.
\EEQ

The $z$-update can be written as
\[
z^{j+1} = \Pi(\mask (x^1)^{j+1} + (u^1)^j, \ldots, \mask (x^K)^{j+1} + 
(u^K)^j),
\]
where $\Pi$ is the projection onto the domain of $g$, \ie,
the constraints $\mask y  =  z^1+\cdots +z^k$.
To simplify notation, let $a^k = \mask (x^k)^{j+1} + (u^k)^{j}$. 
The $z$-update can be written as
\BEAS
(z^k)^{j+1} &=& a^k + (1/K)(\mask y - a^1 - \cdots - a^K) \\
&=& \mask (x^k)^{j+1} + (u^k)^{j} + (1/K)(\mask y - a^1 - \cdots - a^K).
\EEAS

Now consider the $u$-update. 
Plugging in the new $z$-update above, we get
\[
(u^k)^{j+1} = -(1/K)(\mask y - a^1 - \cdots - a^K).
\]
The righthand side does not depend on $k$, which means that all $(u^k)^{j+1}$ 
are the same and can be denoted as $u^{j+1}$. 
(This simplification is not 
unexpected since the original problem has only one dual variable,
which is a vector in $\reals^q$.)
With this simplification, the $u$-update (now for just one scaled dual
variable $u \in \reals^q$) becomes
\[
u^{j+1} = u^j +\frac{1}{K}
\left(\sum_{k=1}^K \mask(x^k)^{j+1} - \mask y\right).
\]

Substituting $u^j$ for $(u^k)^j$ in the $z$-update, we get
\[
(z^k)^{j+1} = \mask (x^k)^{j+1} - u^{j+1}.
\]
Substituting $(z^k)^j = \mask (x^k)^j - u^j$ into the original $x$-update
\eqref{e-x-update1} above, we obtain
\BEAS
(x^k)^{j+1} &=& \argmin_{x^k}\left(\phi_k(x^k) + (\rho/2)\|\mask x^k - 
\mask (x^k)^j + 2 u^j\|_2^2\right) \\
&=& \argmin_{x^k}\left(\phi_k(x^k) + (\rho/2)\|\mask (x^k - 
(x^k)^j + 2 \mask^* u^j)\|_2^2\right) \\
&=& \mprox_{\phi_k}((x^k)^j - 2 \mask^* u^j),
\EEAS
for $k=1, \ldots, K$. (We use \eqref{e-mprox-MM} in the second line.)

We now see that the variables $z^k$ have dropped out, and we arrive at
the final set of ADMM iterations
\BEAS
(x^k)^{j+1} &=& \mprox_{\phi_k}((x^k)^j - 2 \mask^* u^j), \quad k=1, \ldots, K\\
u^{j+1} &=& u^j + \frac{1}{K}
\left(\sum_{k=1}^K \mask(x^k)^{j+1} - \mask y\right).
\EEAS

\end{document}